\documentclass{article}
\PassOptionsToPackage{numbers, compress}{natbib}
\usepackage[final]{neurips_2021}

\usepackage[utf8]{inputenc} % allow utf-8 input
\usepackage[T1]{fontenc}    % use 8-bit T1 fonts
\usepackage{hyperref}       % hyperlinks
\usepackage{url}            % simple URL typesetting
\usepackage{booktabs}       % professional-quality tables
\usepackage{amsfonts}       % blackboard math symbols
\usepackage{nicefrac}       % compact symbols for 1/2, etc.
\usepackage{microtype}      % microtypography
\usepackage{xcolor}         % colors

\usepackage{graphicx}
\usepackage{subcaption}
\usepackage{amsmath}
\usepackage{enumitem}
\usepackage{multirow}
\usepackage{capt-of}
\usepackage{wrapfig}
\usepackage{xcolor}
\usepackage{makecell}
\usepackage{bm}
\usepackage{arydshln}
\usepackage{placeins}
\hypersetup{
    colorlinks = true,
    citecolor = blue,
    linkcolor = red
}

\title{Edge Representation Learning with Hypergraphs}

\author{
  Jaehyeong Jo$^{1*}$,\quad Jinheon Baek$^{1*}$,\quad Seul Lee$^{1}$\thanks{Equal contribution}
  ,\\\textbf{Dongki Kim$^{1}$,\quad Minki Kang$^{1}$,\quad Sung Ju Hwang$^{1,2}$}\\
  KAIST$^{1}$, AITRICS$^{2}$, South Korea \\
  \texttt{harryjo97@kaist.ac.kr, jinheon.baek@kaist.ac.kr,} \\
  \texttt{ellenlee7890@gmail.com, cleverki@kaist.ac.kr,} \\
  \texttt{zzxc1133@kaist.ac.kr, sjhwang82@kaist.ac.kr}
}

\begin{document}

\maketitle

\begin{abstract}
Graph neural networks have recently achieved remarkable success in representing graph-structured data, with rapid progress in both the node embedding and graph pooling methods. Yet, they mostly focus on capturing information from the nodes considering their connectivity, and not much work has been done in representing the \emph{edges}, which are essential components of a graph. However, for tasks such as graph reconstruction and generation, as well as graph classification tasks for which the edges are important for discrimination, accurately representing edges of a given graph is crucial to the success of the graph representation learning. To this end, we propose a novel edge representation learning framework based on \emph{Dual Hypergraph Transformation} (DHT), which transforms the edges of a graph into the nodes of a \emph{hypergraph}. This dual hypergraph construction allows us to apply message-passing techniques for node representations to edges. After obtaining edge representations from the hypergraphs, we then cluster or drop edges to obtain holistic graph-level edge representations. We validate our edge representation learning method with hypergraphs on diverse graph datasets for graph representation and generation performance, on which our method largely outperforms existing graph representation learning methods. Moreover, our edge representation learning and pooling method also largely outperforms state-of-the-art graph pooling methods on graph classification, not only because of its accurate edge representation learning, but also due to its lossless compression of the nodes and removal of irrelevant edges for effective message-passing.\footnote[1]{Code is available at https://github.com/harryjo97/EHGNN}
\end{abstract}
\section{Introduction}
The recent demand in representing graph-structured data, such as molecular, social, and knowledge graphs, has brought remarkable progress in the \emph{Graph Neural Networks} (GNNs)~\cite{GNN/1, GNN/2}.
Early works on GNNs \cite{GCN, GraphSAGE, GIN} aim to accurately represent each node to reflect the graph topology, by transforming, propagating, and aggregating information from their neighborhoods based on message-passing schemes~\cite{MPNN}. 
More recent works focus on learning holistic graph-level representations, by proposing graph pooling techniques that condense the node-level representations into a smaller graph or a single vector. 
While such state-of-the-art node embedding or graph pooling methods have achieved impressive performances on graph-related tasks (e.g., node classification and graph classification), they have largely overlooked the \emph{edges}, which are essential components of a graph.

\begin{figure*}
    \centering
    \includegraphics[width=0.975\linewidth]{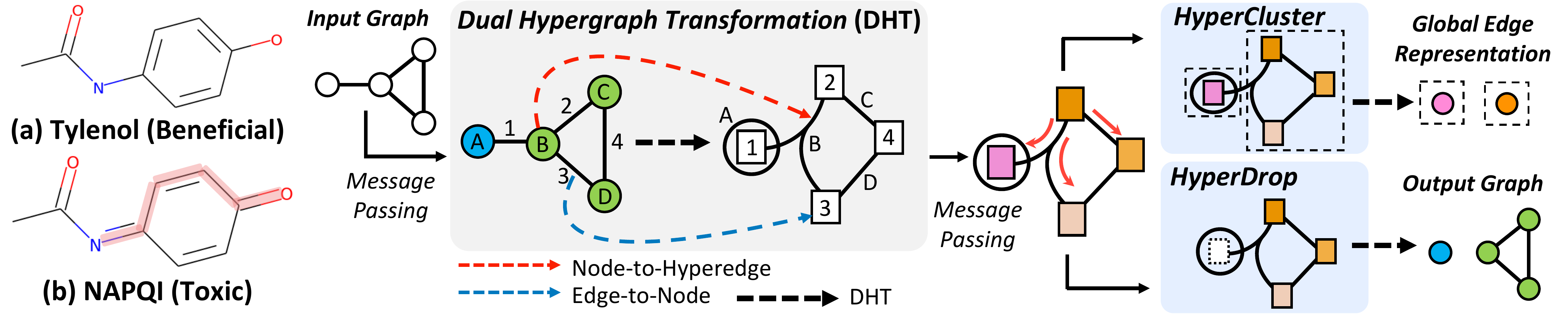}
    \vspace{-0.05in}
    \caption[Caption for LOF]{\small \textbf{(Left):} The two molecular graphs\protect\footnotemark{} have the identical set of nodes, but possess completely different properties due to the difference in edges.
    \textbf{(Right):} An illustration of the proposed edge representation learning framework with two novel edge pooling schemes.
    The grey box in the center describes the proposed \textbf{Dual Hypergraph Transformation}, where the numbers (letters) denote the corresponding edges (nodes) in the graph and nodes (hyperedges) in the hypergraph.
    The two blue boxes in the right illustrate the proposed edge pooling methods, \textbf{HyperCluster} which clusters similar edges,
    and \textbf{HyperDrop} which drops unnecessary edges.
    }
    \label{fig:concept}
    \vspace{-0.2in}
\end{figure*}

Most existing GNNs, including ones that consider categorical edge features~\cite{RelationalGCN, MPNN}, only implicitly capture the edge information in the learned node/graph representations when updating them. While a few of them aim to obtain explicit representations for edges~\cite{EdgeRepresentation/0, EdgeRepresentation/1, EdgeRepresentation/2}, they mostly use them only to augment the node-level representations, and thus suboptimally capture the edge information. This is partly because many benchmark tasks for GNN performance evaluation, such as graph classification, do not require the edge information to be accurately preserved. Thus, on this classification task, simple MLPs without any connectivity information can sometimes outperform GNNs~\cite{benchmark1, benchmark2}. However, for tasks such as graph reconstruction and generation, accurately representing the edges of a graph is crucial to success, as incorrectly reconstructing/generating edges may result in complete failure of the tasks. 
For example, the two molecules (a) and (b) in Figure~\ref{fig:concept} have exactly the same set of nodes and are only different in their edges (bond types), but exhibit extremely different properties.

To overcome such limitations of existing GNN methods in edge representation learning, we propose a simple yet effective scheme to represent the edges. The main challenge of handling edges is the absence or suboptimality of the message-passing scheme for edges. We tackle this challenge by representing the edges as nodes in a \emph{hypergraph}, which is a generalization of a graph that can model higher-order interactions among nodes as one hyperedge can connect an arbitrary number of nodes. Specifically, we propose \emph{Dual Hypergraph Transformation} (DHT) to transform edges of the original graph to nodes of a \emph{hypergraph} (Figure~\ref{fig:concept}), and nodes to hyperedges. This hypergraph-based approach is effective since it allows us to apply any off-the-shelf message-passing schemes designed for node-level representation learning, for learning the representation of the edges of a graph.

However, representing each edge well alone is insufficient in obtaining an accurate representation of the entire graph. Thus we propose two novel graph pooling methods for the hypergraph to obtain compact graph-level edge representations, namely \emph{HyperCluster} and \emph{HyperDrop}. Specifically, for obtaining global edge representations for an entire graph, HyperCluster coarsens similar edges into a single edge under the global graph pooling scheme (see HyperCluster in Figure~\ref{fig:concept}).
On the other hand, HyperDrop drops unnecessary edges from the original graph by calculating pruning scores on the hypergraph (see HyperDrop in Figure~\ref{fig:concept}). 
HyperCluster is more useful for graph reconstruction and generation as it does not result in the removal of any edges, while HyperDrop is more useful for classification as it learns to remove edges that are less useful for graph discrimination.

%%%%%%%%%%%%%%%%%%%%%%%%%%%%%%%%%%%%%%%%%%%%%%%%%%%%%%%%%%%%%
\footnotetext{We depict only the heavy atoms, as conventional preprocessing of molecular graphs drops hydrogen atoms.}
%%%%%%%%%%%%%%%%%%%%%%%%%%%%%%%%%%%%%%%%%%%%%%%%%%%%%%%%%%%%%

We first experimentally validate the effectiveness of the DHT with \emph{HyperCluster}, on the reconstruction of synthetic and molecular graphs. Our method obtains extremely high performance on these tasks, largely outperforming baselines, which shows its effectiveness in accurately representing the edges.
Then, we validate our method on molecular graph generation tasks, and show that it largely outperforms base generation methods, as it allows us to generate molecules with more correct bonds (edges). 
Further, we validate \emph{HyperDrop} on 10 benchmark datasets for graph classification, on which \emph{HyperDrop} outperforms all hierarchical pooling baselines, with larger gains on social graphs, for which the edge features are important. 
Our main contributions are summarized as follows:
\vspace{-0.05in}
\begin{itemize}[itemsep=0.5mm, parsep=1pt, leftmargin=*]
    \item We introduce a \textbf{novel edge representation learning scheme} using \emph{Dual Hypergraph Transformation}, which exploits the dual hypergraph whose nodes are edges of the original graph, on which we can apply off-the-shelf message-passing schemes designed for node-level representation learning.
    \item We propose \textbf{novel edge pooling methods} for graph-level representation learning, namely \emph{HyperCluster} and \emph{HyperDrop}, to overcome the limitations of existing node-based pooling methods.
    \item We validate our methods on \textbf{graph reconstruction, generation, and classification tasks}, on which they largely outperform existing graph representation learning methods.
\end{itemize}

\section{Related Work}

\vspace{-0.05in}
\paragraph{Graph neural networks}
Graph neural networks (GNNs) mostly use the message-passing scheme~\cite{MPNN} to aggregate features from their neighbors.
Particularly, Graph Convolutional Network (GCN)~\cite{GCN} generalizes the convolution operation in the spectral domain of graphs, and updates the representation of each node by applying the shared weights on it and its neighbors' representations. 
Similarly, GraphSAGE~\cite{GraphSAGE} propagates the features of each node's neighbors to itself, based on simple aggregation operations (e.g., mean). 
Graph Attention Network (GAT)~\cite{GAT} considers the relative importance on neighboring nodes with attention, to update each node's representation as the weighted combination of its neighbors'. 
\citet{GIN} show that a simple sum on neighborhood aggregation makes GNNs as powerful as the Weisfeiler-Lehman (WL) test~\cite{WLtest}, which is effective for distinguishing different graphs.
While GNNs have achieved impressive success on graph-related tasks, most of them only focus on learning node-level representations, with less focus on the edges.

\vspace{-0.05in}
\paragraph{Edge-aware graph neural networks}
Some existing works on GNNs consider edge features while updating the node features~\cite{RelationalGCN, Compgcn}, however, they only use the edges as auxiliary information and restrict the representation of edges as the discrete features with categorical values. 
While a few methods~\cite{EdgeRepresentation/0, MPNN, EdgeRepresentation/1, EdgeRepresentation/2} explicitly represent edges by introducing edge-level GNN layers, they use the obtained edge features solely for enhancing node features. Also, existing message-passing schemes for nodes are not directly applicable to edge-level layers, as they are differently designed from the node-level layers, which makes it challenging to combine them with graph pooling methods~\cite{DiffPool} for graph-level representation learning. 
We overcome these limitations by proposing a dual hypergraph transformation scheme, to obtain a hypergraph whose nodes are edges of the original graph, which allows us to apply any message-passing layers designed for nodes to edges.

\vspace{-0.05in}
\paragraph{Graph transformation}
Recently, some works~\cite{linegraph/censnet, linegraph/gcn} propose to transform the original graph into a typical graph structure, to apply graph convolution for learning the edge features. 
Specifically, they construct a line graph~\cite{linegraph}, where the nodes of the line graph correspond to the edges of the original graph, and the nodes of the line graph are connected if the corresponding edges of the original graph share the same endpoint.
However, the line graph transformation has obvious drawbacks: 1) the transformation is not injective, thus two different graphs may be transformed into the same line graph; 2) the transformation is not scalable; 3) node information in the original graph may be lost during the transformation. 
Instead of using such a graph structure, we use hypergraphs, which can model higher-order interactions among nodes by grouping multi-node relationships into a single hyperedge~\cite{berge1973hypergraphs}. 
Using the hypergraph duality~\cite{duality}, edges of the original graph are regarded as the nodes of a hypergraph.
For example, \citet{hypergraphlet} cast a hyperlink prediction task as an instance of node classification from the dual form of the original hypergraph. On the other hand, \citet{hypergraphgrammar} uses the duality to extract useful rules from the hypergraph structures by transforming molecular graphs, for their generation. 
However, none of the existing works exploit the relation between the original graph and the dual hypergraph for edge representation learning.

\vspace{-0.05in}
\paragraph{Graph pooling}
Graph pooling methods aim to learn accurate graph-level representation by compressing a graph into a smaller graph or a vector with pooling operations. The simplest pooling approaches are using \texttt{mean}, \texttt{sum} or \texttt{max} over all node representations~\citep{avg/pooling/1, GIN}. However, they treat all nodes equally, and cannot adaptively adjust the size of graphs for downstream tasks.
More advanced methods, such as node clustering methods, coarsen the graph by clustering similar nodes based on their embeddings~\cite{DiffPool, MincutPool}, whereas the node pruning methods reduce the number of nodes from the graph by dropping unimportant nodes based on their scores~\cite{TopKPool, SAGPool}.
\citet{ASAP} combine both node pruning and clustering approaches, by dropping meaningless clusters after grouping nodes. \citet{gmt} propose to use attention-based operations for considering relationships between clusters.
Note that all of those pooling schemes not only ignore edge representations, but also alter the node set by dropping, clustering, or merging nodes, which result in an inevitable loss of node information.

%%%%%%%%%%%%%%%%%%%%%%%%%%%%%%%%%%%%%%%%%%%%%%%%%%%%%%%%%%%%%
\begin{figure}[!t]
    \centering
    \includegraphics[width=0.95\linewidth]{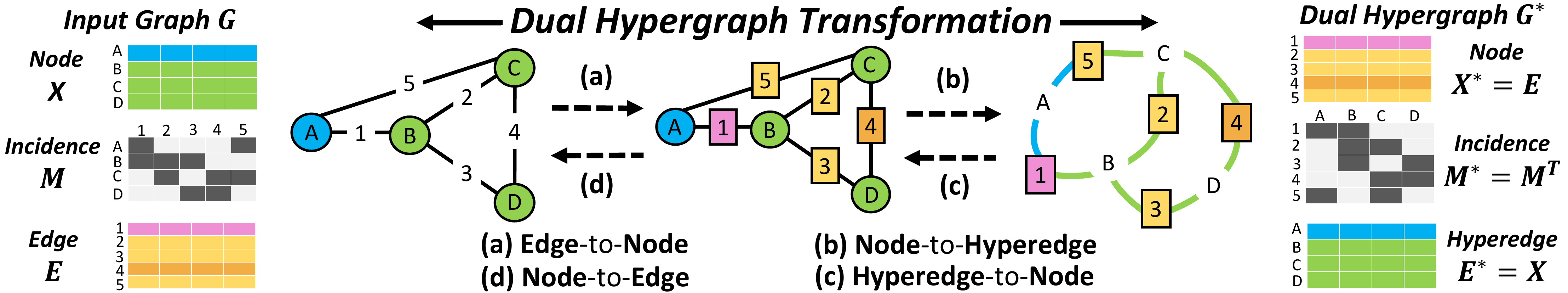}
    %\vspace{-0.05in}
    \caption{\small \textbf{Dual Hypergraph Transformation.} 
    Illustration of the proposed graph-to-hypergraph transformation. 
    }
    \vspace{-0.15in}
    \label{fig:DHT}
\end{figure}
%%%%%%%%%%%%%%%%%%%%%%%%%%%%%%%%%%%%%%%%%%%%%%%%%%%%%%%%%%%%%

\section{Edge Representation Learning with Hypergraphs}
In this section, we first introduce our novel edge representation learning framework with dual hypergraphs, which we refer to as \textit{Edge HyperGraph Neural Network} (EHGNN), and then propose two novel edge pooling schemes for holistic graph-level representation learning: \emph{HyperCluster} and \emph{HyperDrop}.
We begin with the descriptions of graph neural networks for node representation learning.

\paragraph{Graph neural networks}
A graph $G$ with $n$ nodes and $m$ edges, is defined by its node features $\bm{X}\in\mathbb{R}^{n\times d}$, edge features $\bm{E}\in\mathbb{R}^{m\times d'}$, and the connectivity among the nodes represented by an adjacency matrix $\bm{A}\in\mathbb{R}^{n\times{n}}$. 
Here, $d$ and $d'$ are the dimensions of node and edge features, respectively. Then, given a graph, the goal of a \emph{Graph Neural Network} (GNN) is to learn the node-level representation with message-passing between neighboring nodes~\cite{MPNN} as follows:
\begin{equation}
    \bm{X}^{(l+1)}_v
    = \text{UPDATE}\left( \bm{X}^{(l)}_v, \text{AGGREGATE}\left(\left\{  \bm{X}^{(l)}_u: \forall u\in\mathcal{N}(v; \bm{A}) \right\}\right) \right),
    \label{eq:mp}
\end{equation}
where $\bm{X}^{(l)}$ is the node features at $l$-th layer, AGGREGATE is the function that aggregates messages from a set of neighboring nodes of the node $v$, UPDATE is the function that updates the representation of the node $v$ from the aggregated messages, and $\mathcal{N}(v;\bm{A})$ is the set of neighboring nodes for the node $v$, obtained from the adjacency matrix $\bm{A}$.
Such message-passing schemes can incorporate the graph topology into each node by updating its representation with the representation of its neighbors.

\subsection{Edge representation learning with dual hypergraph transformation}
\label{subsec:framework}

\paragraph{Edge representation learning}
To reflect the edge information on message-passing, some works on GNNs first obtain the categorical edge features between nodes, and then use them on the AGGREGATE function in equation~\ref{eq:mp}, by adding or multiplying the edge features to the neighboring node's features~\cite{MPNN, RelationalGCN} (see Appendix~\ref{sup/edge/aware} for more details).
Similarly, few recent works aim to obtain explicit edge representations, but only to use them as auxiliary information to augment the node features, by adding or multiplying edge features to them~\cite{EdgeRepresentation/1, EdgeRepresentation/2}.
Thus, existing works only implicitly capture the edge information in the learned node representations. 
Although this might be sufficient for most benchmark graph classification tasks, many real-world tasks of graphs (e.g., graph reconstruction and generation) further require the edges to be accurately represented as the information on edges could largely affect the task performance.

%%%%%%%%%%%%%%%%%%%%%%%%%%%%%%%%%%%%%%%%%%%%%
\begin{wraptable}{h}{0.45\textwidth}
\small
\centering
\vspace{-0.165in}
\caption{\small Transformation and message-passing complexities of edge-aware GNNs, line graph, and our EHGNN for the star graph, in which one hub node is connected to $n$ other nodes.}
\vspace{-0.05in}
\resizebox{0.45\textwidth}{!}{
    \renewcommand{\arraystretch}{0.5}
    \renewcommand{\tabcolsep}{1.3mm}
    \begin{tabular}{lccc}
    \toprule
    \multirow{2}{*}{\bf Models} & \multicolumn{2}{c}{\textbf{Complexity}} \\
    & {\bf Transformation} & {\bf Message-passing} \\
    \midrule
    Edge-aware GNNs & $\mathcal{O}(n^2)$ & $\mathcal{O}(n)$ \\
    Line graph & $\mathcal{O}(n^2)$ & $\mathcal{O}(n^2)$ \\
    \midrule
    EHGNN (Ours) & $\mathcal{O}(n)$ & $\mathcal{O}(n)$ \\
    \bottomrule
    \end{tabular}
}
\vspace{-0.125in}
\label{tbl:transformation}
\end{wraptable}
%%%%%%%%%%%%%%%%%%%%%%%%%%%%%%%%%%%%%%%%%%%%%

Even worse, to define a message-passing function for edge representation learning, existing works propose to additionally create the adjacency matrix for edges, either by defining the edge neighborhood structure~\cite{EdgeRepresentation/2} or using the line graph transformation~\cite{EdgeRepresentation/0}.
However, these are highly suboptimal as obtaining the adjacency of edges
requires $\mathcal{O}(n^2)$ time complexity (see Appendix~\ref{sup/edge/complexity} for detailed descriptions), as shown in Table~\ref{tbl:transformation}.
This is the main obstacle for directly applying existing message-passing schemes for nodes to edges.
To this end, we propose a simple yet effective method to represent the edges of a graph, using a hypergraph.

\paragraph{Hypergraph}
A \emph{hypergraph} is a generalization of a graph that can model graph-structured data with higher-order interactions among nodes, wherein a single hyperedge connects an arbitrary number of nodes, unlike in conventional graphs where an edge can only connect two nodes.
For example, in Figure~\ref{fig:DHT}, the hyperedge $B$ defines the relation among three different nodes. To denote such higher-order relations among arbitrary number of nodes defined by a hyperedge, we use an \emph{incidence matrix} $\bm{M}\in\{0,1\}^{n\times m}$, which represents the interaction between $n$ nodes and $m$ hyperedges, instead of using an adjacency matrix $\bm{A}\in\{0,1\}^{n\times n}$ that only considers interactions among $n$ nodes.
Each entry in the incidence matrix indicates whether the node is incident to the hyperedge.
We can formally define a hyperagraph $G^{\ast}$ with $n$ nodes and $m$ hyperedges, as a triplet of three components $G^{\ast} = (\bm{X}^{\ast}, \bm{M}^{\ast}, \bm{E}^{\ast})$,
where $\bm{X}^{\ast}\in\mathbb{R}^{n\times d}$ is the node features, $\bm{E}^{\ast}\in\mathbb{R}^{m\times d'}$ is the hyperedge features, and $\bm{M}^{\ast}\in\{0,1\}^{n\times m}$ is the incidence matrix of the hypergraph. We can also represent conventional graphs in the form of a hypergraph, $G = (\bm{X}, \bm{M}, \bm{E})$, in which a hyperedge in the incidence matrix $\bm{M}$ is associated with only two nodes. 
In the following paragraph, we will describe how to transform the edges of a graph into nodes of a hypergraph, for edge representation learning.

\paragraph{Dual Hypergraph Transformation} 
If we can change the role of the nodes and edges of the graph with a shared connectivity pattern across the nodes and edges, while accurately preserving their information, then we can use any node-based message-passing schemes for learning edges.
To achieve this, inspired by the hypergraph duality~\cite{berge1973hypergraphs, duality}, we propose to transform an edge of the original graph into a node of a hypergraph, and a node of the original graph into a hyperedge of the same hypergraph. We refer to this graph-to-hypergraph transformation as \emph{Dual Hypergraph Transformation} (DHT) (see Figure~\ref{fig:DHT}).
To be more precise, during the transformation, we interchange the structural role of nodes and edges from the given graph, obtaining the incidence matrix for the new dual hypergraph simply by \emph{transposing} the incidence matrix of the original graph (see the incidence matrix in Figure~\ref{fig:DHT}).
Along with the structural transformation through the incidence matrix, the DHT naturally interchanges node and edge features across $G$ and $G^{\ast}$ (see the feature matrices in Figure~\ref{fig:DHT}).
Formally, given a triplet representation of a graph, DHT is defined as the following transformation:
\begin{equation}
    \bm{DHT} \;:\; 
    G = \big( \bm{X}, \bm{M}, \bm{E} \big) \;\mapsto\; G^{\ast} = \big(\bm{E},\bm{M}^T,\bm{X} \big),
    \label{eq:DHT}
\end{equation}
where we refer to the transformed $G^{\ast}$ as the \emph{dual hypergraph} of the input graph $G$. Since the dual hypergraph $G^{\ast}=(\bm{E}, \bm{M}^T, \bm{X})$ retains all the information of the original graph, 
we can recover the original graph from the dual hypergraph with the same DHT operation as follows:
\begin{equation}
    \bm{DHT} \;:\; 
    G^{\ast} = \big( \bm{E}, \bm{M}^T, \bm{X} \big) \;\mapsto\; G = \big( \bm{X}, \bm{M}, \bm{E} \big).
    \label{eq:DHTinv}
\end{equation}
This implies that DHT is a \emph{bijective transformation}. DHT is simple to implement, does not incur the loss of any features or topological information of the input graph, and does not require additional memory for feature representations. 
Moreover, DHT can be sparsely implemented using the edge list, which is the sparse form of the adjacency matrix, by only reshaping the edge list of the original graph into the hyperedge list of the dual hypergraph (see Appendix~\ref{sup/edge/sparse} for details), which is highly efficient in terms of time and memory.
Thanks to DHT, we define the message-passing between edges of the original graph as the message-passing between nodes of its dual hypergraph.

\paragraph{Message-passing on the dual hypergraph for edge representation learning}
After transforming the original graph into its corresponding dual hypergraph using DHT, we can perform the message-passing between edges of the input graph, by performing the message-passing between nodes of its dual hypergraph $G^{\ast}=(\bm{E},\bm{M}^T,\bm{X})$, which is formally denoted as follows:
\begin{equation}
    \bm{E}^{(l+1)}_e = \text{UPDATE}\left(
    \bm{E}^{(l)}_e, \text{AGGREGATE}\left(
    \left\{ \bm{E}^{(l)}_f:\forall f\in\mathcal{N}(e; \bm{M}^T) \right\}
    \right)
    \right),
    \label{eq:hypermp}
\end{equation}
where $\bm{E}^{(l)}$ is the node features of $G^{\ast}$ at $l$-th layer, the \text{AGGREGATE} function summarizes the neighboring messages of the node $e$ of the dual hypergraph $G^{\ast}$, and the $\text{UPDATE}$ function updates the representation of the node $e$ from the aggregated messages.
Here $\mathcal{N}(e;\bm{M}^T)$ is the neighboring node set of the node $e$ in $G^{\ast}$, which we obtain using the incidence matrix $\bm{M}^T$ of $G^{\ast}$. 
Furthermore, instead of using the dense incidence matrix, we can sparsely implement the message-passing on the dual hypergraph with the hyperedge list, from which the complexity of message-passing on the dual hypergraph reduces to $\mathcal{O}(m)$, which is equal to the complexity of message-passing between nodes on the original graph (See Appendix~\ref{sup/edge/complexity} for details).
Note that, since the form of equation~\ref{eq:hypermp} is the same as the form of equation~\ref{eq:mp}, we can use any graph neural networks which realize the message-passing operation in equation~\ref{eq:mp}, such as GCN~\cite{GCN}, GAT~\cite{GAT}, GraphSAGE~\cite{GraphSAGE}, and GIN~\cite{GIN}, for equation~\ref{eq:hypermp}.
In other words, to learn the edge representations $\bm{E}$ of the original graph, we do not require any specially designed layers, but simply need to perform DHT to directly apply existing off-the-shelf message-passing schemes to the transformed dual hypergraph. 

To simplify, we summarize the equation~\ref{eq:mp} as follows: $\bm{X}^{(l+1)} = \text{GNN}\left(\bm{X}^{(l)}, \bm{M}, \bm{E}^{(l)}\right)$, and the equation~\ref{eq:hypermp} as follows: $\bm{E}^{(l+1)} = \text{GNN}\left(\bm{E}^{(l)}, \bm{M}^T, \bm{X}^{(l)}\right) = \text{EHGNN}\left(\bm{X}^{(l)}, \bm{M}, \bm{E}^{(l)}\right)$, where EHGNN indicates our edge representation learning framework using DHT. 
After updating the edge features $\bm{E}^{(L)}$ with EHGNN, $\bm{E}^{(L)}$ is returned to the original graph by applying DHT on the dual hypergraph $G^{\ast}$. Then, the remaining step is how to make use of these edge-wise representations to accurately represent the edges of the entire graph, which we describe in the next subsection.

\subsection{Graph-level edge representation learning with edge pooling}
\label{subsec:HyperPool}

Existing graph pooling methods do not explicitly represent edges. To overcome this limitation, we propose two novel edge pooling schemes: \emph{HyperCluster} and \emph{HyperDrop}.

\paragraph{Graph pooling}
The goal of graph pooling is to learn a holistic representation of the entire graph. 
The most straightforward approach for this is to aggregate all the node features with \texttt{mean} or \texttt{sum} operations~\cite{avg/pooling/1, GIN}, but they treat all nodes equally without consideration of which nodes are important for the given task. To tackle this limitation, recent graph pooling methods propose to either cluster and coarsen nodes~\cite{DiffPool, MincutPool} or drop unnecessary nodes~\cite{TopKPool, SAGPool}. While they yield improved performances on graph classification tasks, they suffer from an obvious drawback: inevitable loss of both node and edge information. 
The node information is lost as nodes are dropped and coarsened, and the edge information is lost as edges for the dropped nodes or internal edges for the coarsened nodes are removed. To overcome this limitation, we propose a graph-level edge representation learning scheme.

\paragraph{HyperCluster}
We first introduce \emph{HyperCluster}, which is a novel edge clustering method to coarsen similar edges into a single edge, for obtaining the global edge representation.
Generally, a clustering scheme for nodes of the graph~\cite{DiffPool, MincutPool} is defined as follows:
\begin{equation}
    \bm{X}^{pool} = \bm{C}^{T}\bm{X'}, \;\;
    \bm{M}^{pool} = \bm{C}^{T}\bm{M},
\end{equation}
where $\bm{X}^{pool} \in \mathbb{R}^{n_{pool} \times d}$ and $\bm{M}^{pool} \in \mathbb{R}^{n_{pool} \times m}$ denote the pooled representations, $\bm{X'}=\text{GNN}\left( \bm{X}, \bm{M}, \bm{E}\right)\in \mathbb{R}^{n \times d}$ is the updated node features, and $\bm{C} \in \mathbb{R}^{n \times n_{pool}}$ is the cluster assignment matrix that is generated from the $\bm{X'}$. 
Following this approach, the proposed HyperCluster clusters similar edges into a single edge, by clustering nodes of the dual hypergraph obtained from the original graph via DHT. In other words, we first obtain the node representation of the dual hypergraph $\bm{E'} = \text{EHGNN}\left(\bm{X}, \bm{M}, \bm{E}\right) \in \mathbb{R}^{m \times d'}$, and then cluster the nodes of the dual hypergraph as follows:
\begin{equation}
    \bm{E}^{pool} = \bm{C}^{T} \bm{E'} \,,\;\;
    (\bm{M}^{pool})^{T} = \bm{C}^{T}\bm{M}^{T}
\end{equation}
where $\bm{E}^{pool} \in \mathbb{R}^{m_{pool} \times d'}$ and $\bm{M}^{pool} \in \mathbb{R}^{n\times m_{pool}}$ denote the pooled edge representation and the incidence matrix of the input graph respectively, and $\bm{C} \in \mathbb{R}^{m \times m_{pool}}$ is the cluster assignment matrix generated from the input edge features $\bm{E'}$. 
Since HyperCluster coarsens the edges rather than dropping them, this edge pooling method is more appropriate for tasks such as graph reconstruction.

\paragraph{HyperDrop}
We propose another edge pooling scheme, \emph{HyperDrop}, which drops unnecessary edges to identify task-relevant edges, while performing lossless compression of nodes.
Conventional node drop methods~\cite{TopKPool, SAGPool} remove less relevant nodes based on their scores, as follows:
\begin{equation}
    \bm{X}^{pool} = \bm{X}_{idx} \, , \;
    \bm{M}^{pool} = \bm{M}_{idx}  \; ; \;\; 
    idx = \text{top}_k(\text{score}(\bm{X})),
\end{equation}
where $idx$ is the row-wise (i.e., node-wise) indexing vector, $\text{score}(\cdot)$ computes the score of each node with learnable parameters, and $\text{top}_k(\cdot)$ selects the top $k$ elements in terms of the score.
However, this approach results in the inevitable loss of node information, as it drops nodes.
Thus, we propose to coarsen the graph by dropping edges instead of nodes, exploiting edge representations obtained from our EHGNN.
\emph{HyperDrop} selects the top-ranked edges of the original graph, by selecting the top-ranked nodes of the dual hypergraph. The pooling procedure for HyperDrop is as follows:
\begin{equation}
    \bm{E}^{pool} =  \bm{E}_{idx} \, , \;
    (\bm{M}^{pool})^T = (\bm{M}^T)_{idx} \; ; \; \;
    idx = \text{top}_k\left(\text{score}(\bm{E}\right)).
\end{equation}
Then, we can obtain the pooled graph $G^{pool}=(\bm{X},\bm{M}^{pool},\bm{E}^{pool})$ by applying DHT to the pooled dual hypergraph.
HyperDrop is most suitable for graph classification tasks, as it identifies discriminative edges for the given task. Since HyperDrop preserves the nodes intact, it can also be used for node-level classification tasks, which is impossible with exiting graph pooling methods that modify nodes. 
In another point of view, the proposed HyperDrop can be further considered as a learnable graph rewiring operation, which optimizes the graph for the given task by deciding whether to drop or maintain the nodes.
Finally, a notable advantage of such HyperDrop is that it alleviates the over-smoothing problem in deep GNNs~\cite{oversmoothing} (i.e., the features of all nodes converge to the same values when stacking a large number of GNN layers).
As HyperDrop \emph{learns} to remove unnecessary edges, the message-passing only happens across relevant nodes, which alleviates over-smoothing.

\section{Experiments}
\vspace{-0.05in}
We experimentally validate the effectiveness of EHGNN coupled with either HyperCluster or HyperDrop on four different tasks: graph reconstruction, generation, classification, and node classification.

\subsection{Graph reconstruction}
Accurately reconstructing the edge features is crucial for graph reconstruction tasks, and thus we validate the efficacy of our method on graph reconstruction tasks first.

\paragraph{Experimental setup}
We first validate our EHGNN with HyperCluster on the \emph{edge reconstruction} tasks, where the goal is to reconstruct the edge features from their compressed representations.
Then, we evaluate our method on the graph reconstruction tasks to validate the effectiveness of ours in holistic graph-level learning.
We start with edge reconstruction of a synthetic two-moon graph, where node features (coordinates) are fixed and edge features are colors.
For edge and graph reconstruction of real-world graphs, we use the ZINC dataset~\cite{ZINC} that consists of 12K molecular graphs~\cite{benchmarkingGNN}, where node features are atom types and edge features are bond types.
We use accuracy, validity, and exact match as evaluation metrics. 
For more details, please see Appendix~\ref{sup/exp/recon}.

%%%%%%%%%%%%%%%%%%%%%%%%%%%%%%%%%%%%%%%%%%%%%%%%%%%%%%%%%%%%%
\begin{figure}[t!]
    \begin{minipage}{0.58\linewidth}
        \centering
        \caption{\small \textbf{Edge reconstruction results} on the ZINC molecule dataset by varying the pooling ratio. Solid lines denote the mean, and shaded areas denote the standard deviation of 5 runs.
        }
        \vspace{-0.075in}
        \includegraphics[width=1\linewidth]{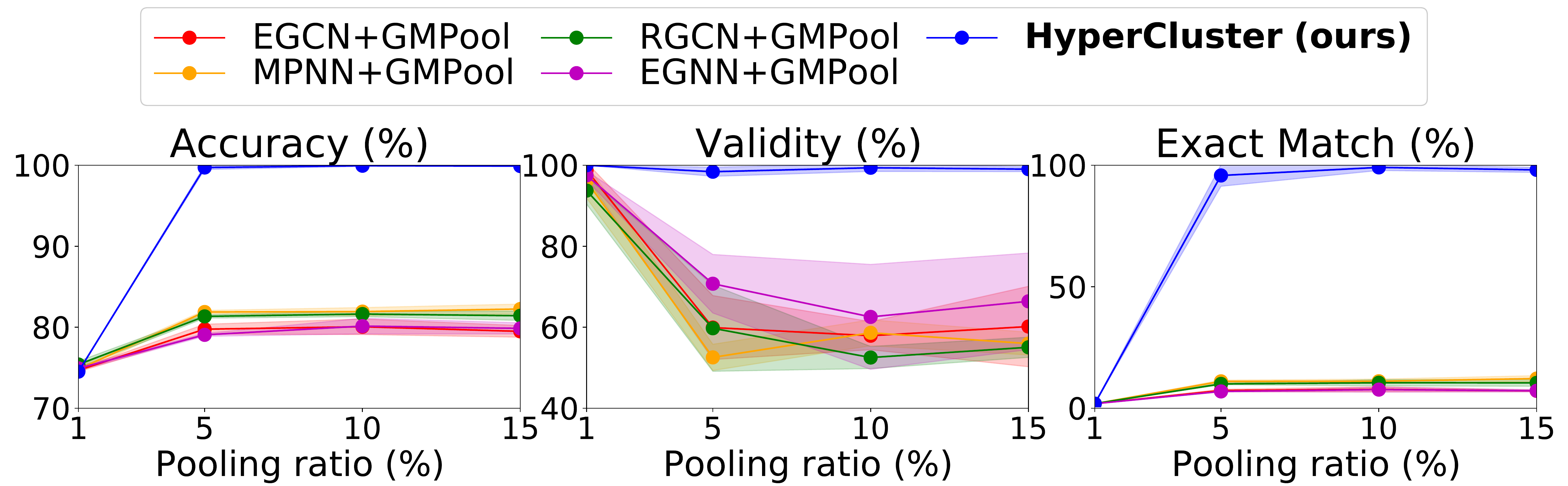}
        \label{fig:recon_edge}
    \end{minipage}
    \hfill
    \begin{minipage}{0.4\linewidth}
        \centering
        \caption{\small \textbf{Edge reconstruction results} of the synthetic two-moon graph. The edge features are represented by colors.
        }
        \vspace{-0.075in}
        \includegraphics[width=1\linewidth]{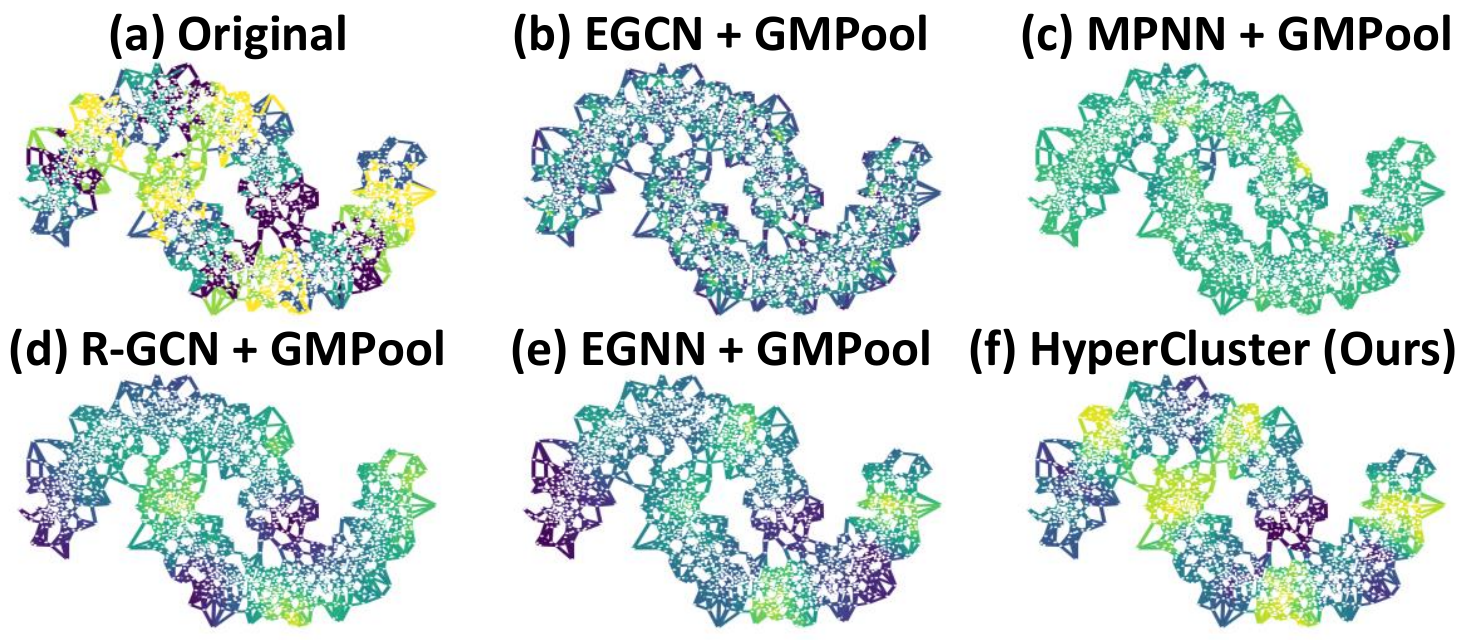}
        \label{fig:recon_grid}
    \end{minipage}
    \vspace{-0.075in}
\end{figure}

\begin{figure}[t!]
    \vspace{-0.075in}
    \begin{minipage}{0.58\linewidth}
        \centering
        \vspace{-0.1in}
        \includegraphics[width=1\linewidth]{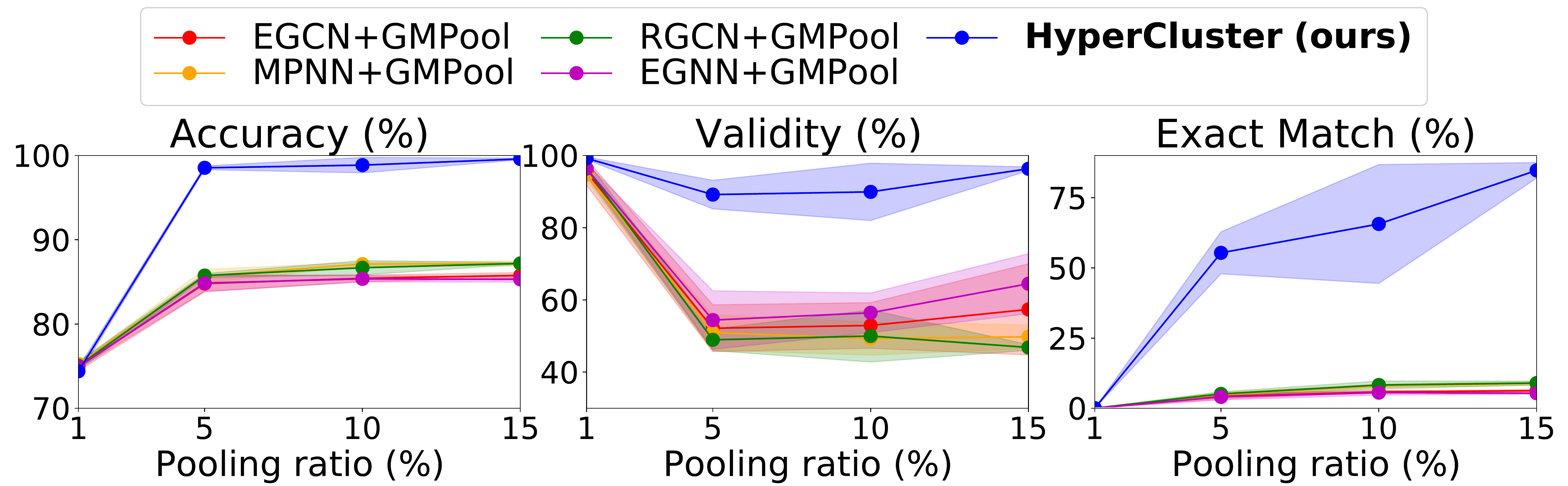}
        \vspace{-0.2in}
        \caption{\small \textbf{Graph reconstruction results} on the ZINC molecule dataset by varying the pooling ratio. Solid lines denote the mean, and shaded areas denote the standard deviation of 5 runs. 
        }
        \label{fig:recon}
    \end{minipage}
    \hfill
    \begin{minipage}{0.4\linewidth}
        \centering
        \vspace{-0.1in}
        \centerline{
        \includegraphics[width=1\linewidth]{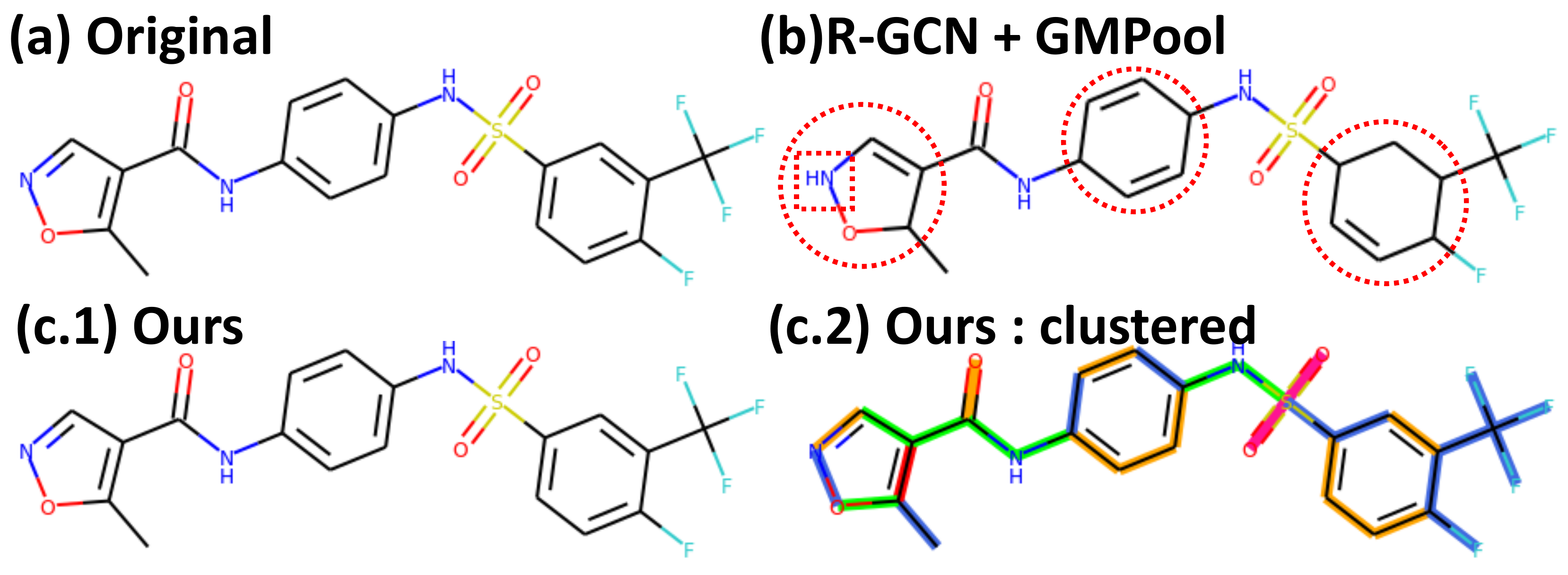}}
        \vspace{-0.12in}
        \caption{\small\textbf{Graph reconstruction examples.} Red dashed circles and squares indicate the incorrectly predicted edges and nodes, respectively.
        (c.2) shows the assigned clusters of edges as colors using our method.
        }
        \label{fig:mol}
    \end{minipage}
    \vspace{-0.15in}
\end{figure}
%%%%%%%%%%%%%%%%%%%%%%%%%%%%%%%%%%%%%%%%%%%%%%%%%%%%%%%%%%%%%

\paragraph{Implementation details and baselines}
We compare the proposed EHGNN framework against edge-aware GNNs, namely EGCN~\cite{OGB}, MPNN~\cite{MPNN}, R-GCN~\cite{RelationalGCN}, and EGNN~\cite{EdgeRepresentation/1}, which use the edge features as auxiliary information for updating node features.
We further combine them with an existing graph pooling method, namely GMPool~\cite{gmt}, to obtain a graph-level edge representation for a given graph. In contrast, for our method, we first obtain edge representations with EHGNN, using GCN~\cite{GCN} as the message-passing function, and then coarsen the edge-wise representations using HyperCluster, whose cluster assignment matrices are obtained using GMPool~\cite{gmt}.
For node reconstruction, we set message-passing to GCN and graph pooling to GMPool~\cite{gmt} for all models.
We provide further details of the baselines and our model in Appendix~\ref{sup/exp/recon}.

\paragraph{Edge reconstruction results}
Figure~\ref{fig:recon_grid} shows the original two-moon graph and edge-reconstructed graphs, where edge features are represented as colors, exhibiting clustered patterns. 
The baselines fail to reconstruct the edge colors, since they implicitly learn edge representations by using edge features as auxiliary information to update nodes, hence mixing the colors of the neighboring edges.
On the other hand, our method distinguishes each edge cluster, which shows that our method can capture meaningful edge information by clustering similar edges.
Moreover, as shown in Figure~\ref{fig:recon_edge}, our model obtains significantly higher performance over all baselines on the edge reconstruction task of molecular graphs, in all evaluation metrics. The performance gain of our method over baselines is notably large in exact match, which demonstrates that explicit learning of edge representation is essential for the accurate encoding of the edge information.

%%%%%%%%%%%%%%%%%%%%%%%%%%%%%%%%%%%%%%%%%%%%%%%%%%%%%%%%%%%%%
\begin{figure}[!t]
    \begin{minipage}{0.50\linewidth}
        \centering
        \begin{subfigure}{0.49\textwidth}
            \includegraphics[width=\linewidth]{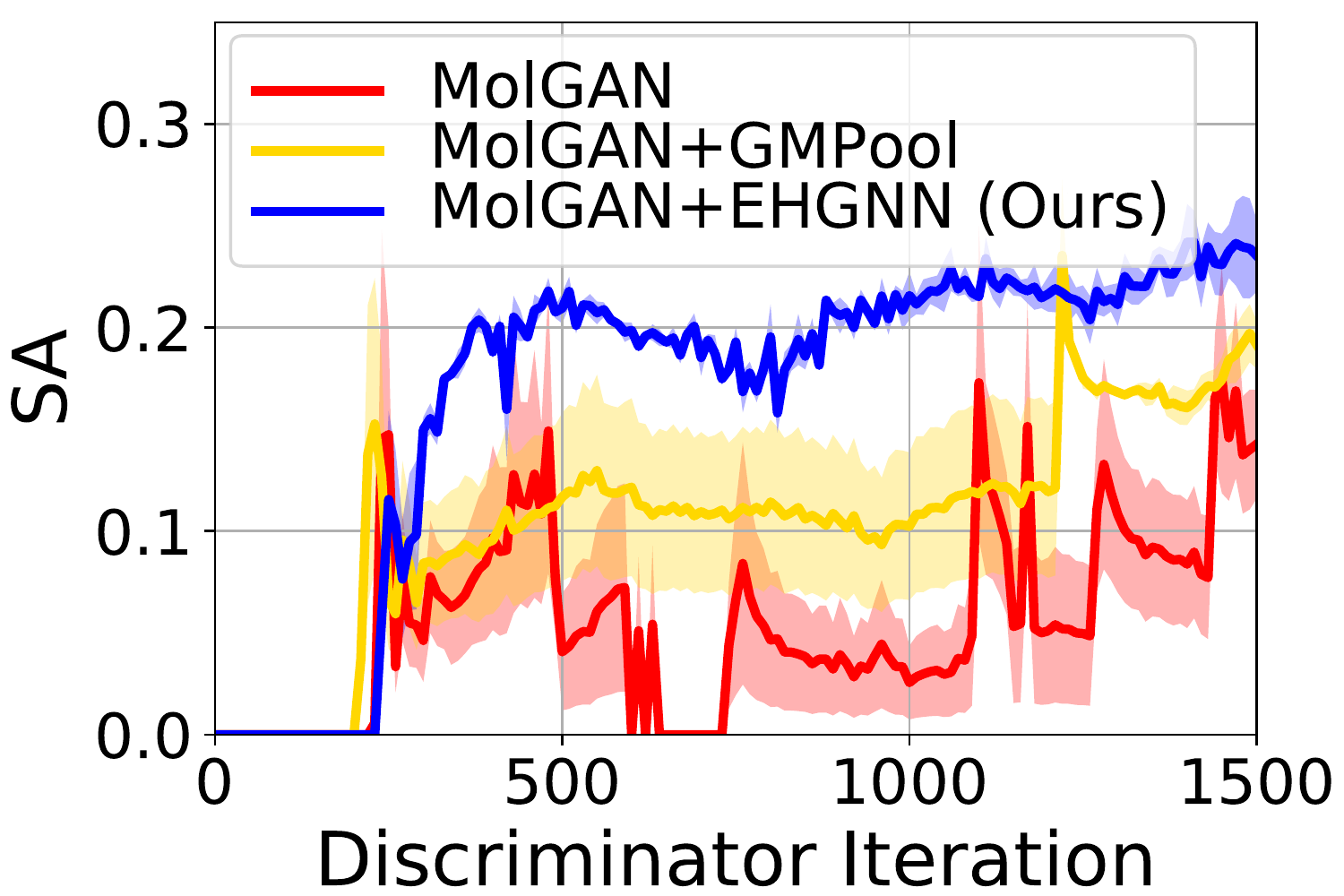}
            %\label{fig:molgan_sa}
        \end{subfigure}
        \hspace*{\fill}%
        \begin{subfigure}{0.49\textwidth}
            \includegraphics[width=\linewidth]{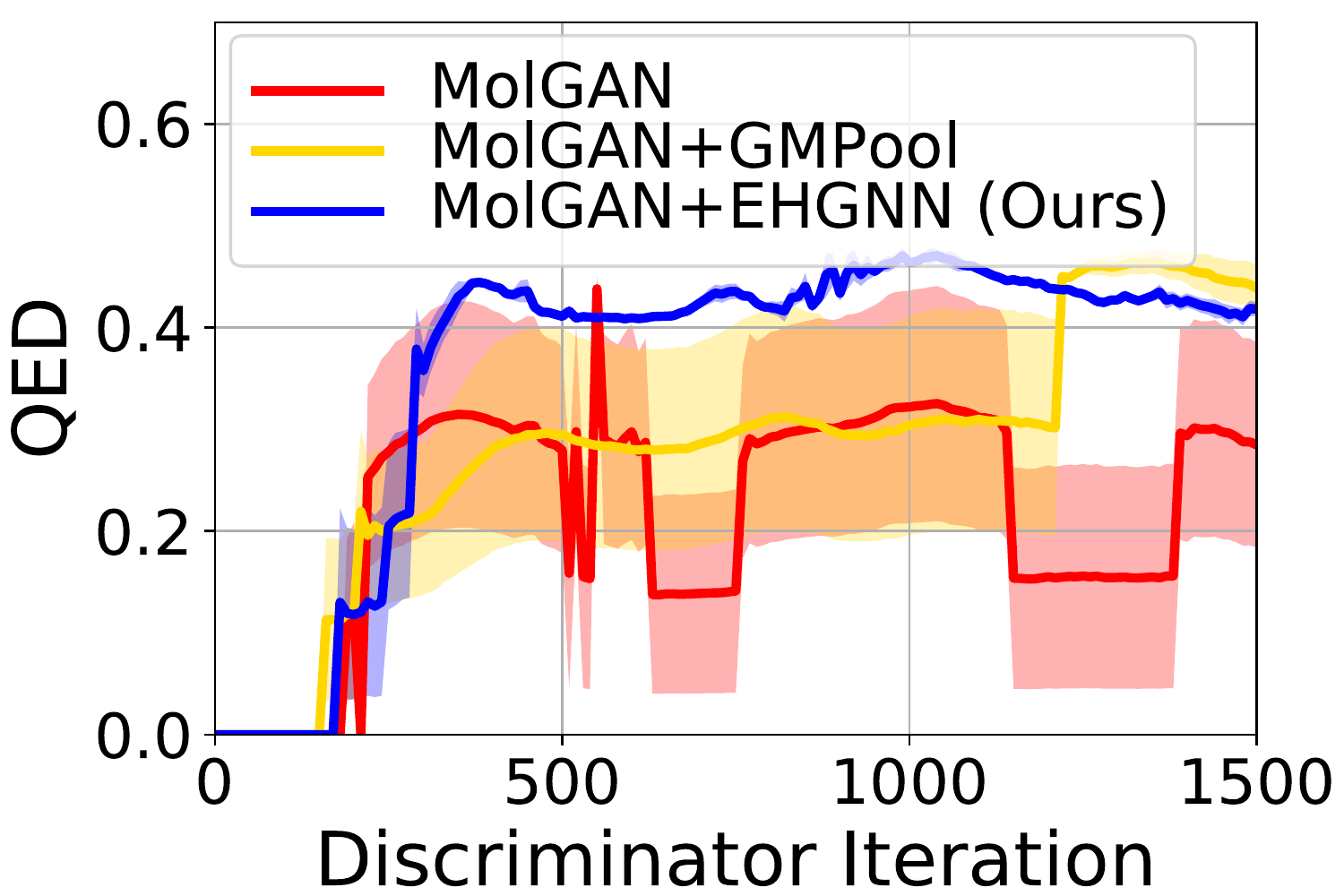}
            %\label{fig:molgan_qed}
        \end{subfigure}
        \vspace{-0.08in}
        \caption{\small \textbf{Graph generation results on MolGAN.} Solid lines denote the mean, and shaded areas denote the standard deviation of 3 different runs.}
        \label{fig:molgan}
    \end{minipage}
    \hfill
    \begin{minipage}{0.48\linewidth}
        \centering
        \resizebox{1\textwidth}{!}{
        \renewcommand{\arraystretch}{1.2}
        \renewcommand{\tabcolsep}{1.0mm}
        \begin{tabular}{lcccc}
        \toprule
            \textbf{Dataset} & \textbf{Metrics} &\textbf{MARS}~\cite{MARS} & \textbf{MARS + EHGNN (Ours)} \\
        \midrule
            \multirow{4}{*}{ZINC15} & Success Rate & 59.53 $\pm$ 2.11 & \textbf{64.30} $\pm$ 1.54 \\
            & QED ($\ge$ 0.67) & 95.71 $\pm$ 0.09 & \textbf{96.36} $\pm$ 0.49 \\
            & GSK3$\beta$ ($\ge$ 0.6) & 86.52 $\pm$ 1.67 & \textbf{90.63} $\pm$ 2.57 \\
            & JNK3 ($\ge$ 0.6) & 71.52 $\pm$ 4.15 & \textbf{73.60} $\pm$ 1.29  \\
        \bottomrule
        \end{tabular}}
        \vspace{-0.06in}
        \captionof{table}{\small \textbf{Graph generation results on MARS.} The results are the mean and standard deviation of 3 different runs. Best performance and its comparable results ($p>0.05$) from the t-test are highlighted in bold.}
        \label{tab:generation_MARS}
    \end{minipage}
    \vspace{-0.1in}
\end{figure}
%%%%%%%%%%%%%%%%%%%%%%%%%%%%%%%%%%%%%%%%%%%%%%%%%%%%%%%%%%%%%

\paragraph{Graph reconstruction results}
To verify the effectiveness of learning accurate edge representations for reconstructing both the node and edge features, we now validate our method on the molecular graphs in Figure~\ref{fig:recon}.
Combining our edge representation learning method (EHGNN + HyperCluster) with the existing node representation learning method (GCN + GMPool) yields incomparably high reconstruction performance compared to the baselines in exact match, which demonstrates that learning accurate edge representation, as well as node representation, is crucial to the success of the graph representation learning methods on graph reconstruction.

\paragraph{Qualitative analysis}
We visualize the original and reconstructed molecular graphs in Figure~\ref{fig:mol}. 
As shown in Figure~\ref{fig:mol} (b), the baseline cannot reconstruct the ring structures of the molecule, whereas our method perfectly reconstructs the rings as well as the atom types.
The generated edge clusters in Figure~\ref{fig:mol} (c.2) further show that our method captures the detailed substructures of the molecule, as we can see the cluster patterns of hexagonal and pentagonal rings. More reconstruction examples of molecular graphs are shown in Figure~\ref{fig:mol_additional}.

\begin{wrapfigure}{!t}{0.36\linewidth}
    \vspace{-0.15in}
    \centering
    \centerline{\includegraphics[width=0.875\linewidth]{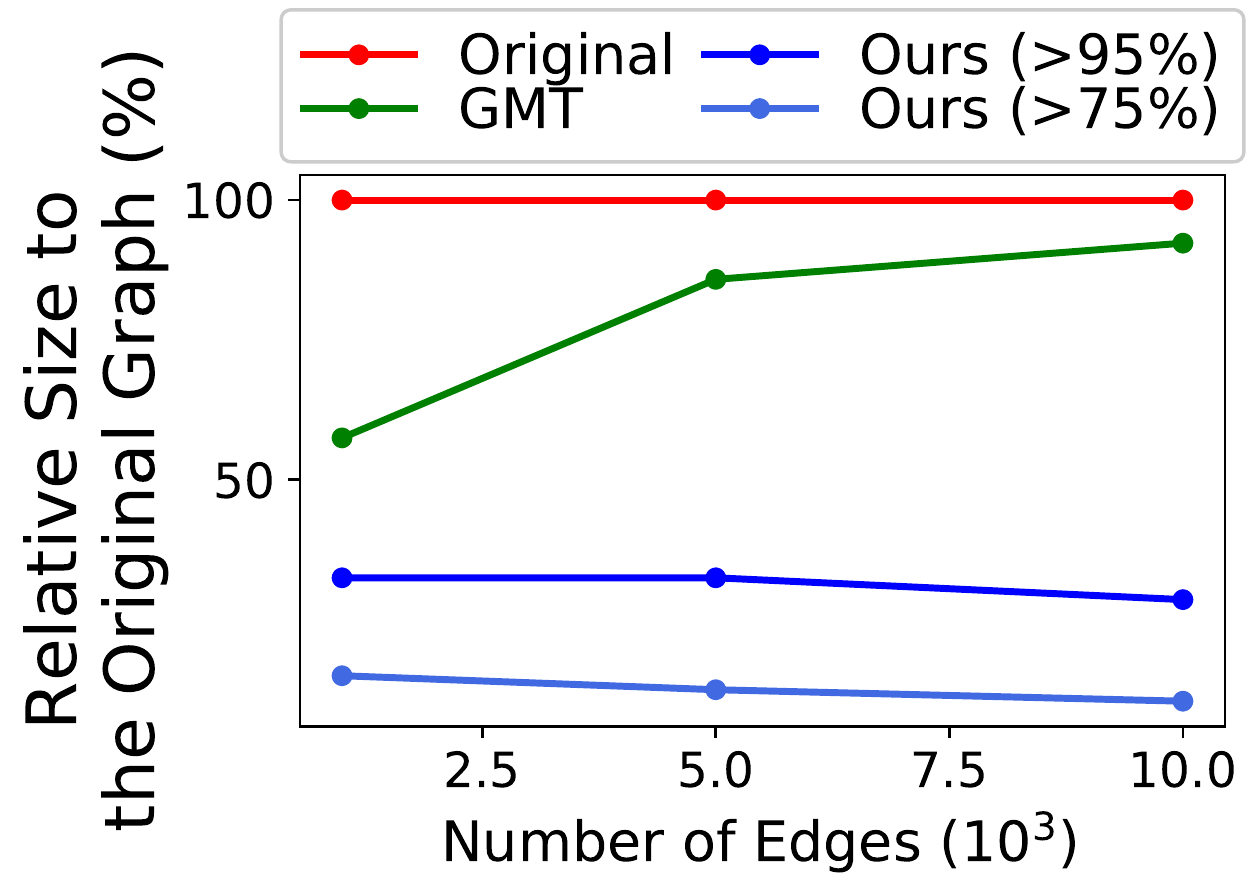}}
    \vspace{-0.1in}
    \caption{\small \textbf{Graph compression results.} For ours, we report the relative size to the original graph where the edge reconstruction accuracy is higher than 95\% and 75\%, respectively.}
    \label{fig:memory}
    \vspace{-0.2in}
\end{wrapfigure}

\paragraph{Graph compression}
To validate the effectiveness of our method in large and dense graph compression, we further apply EHGNN with HyperCluster to the Erdos-Renyi random graph~\cite{randomgraph} having six discrete edge features, where the number of nodes is fixed to $10^3$ while the number of edges increases from $10^3$ to $10^4$. In Figure~\ref{fig:memory}, 
we report the relative memory size of the compressed graph after pooling the features, against the size of the original graph. We compare our method which compresses both the nodes and edges, against the node pooling method, namely GMT~\cite{gmt}.
As the number of edges increases, we observe that compressing only the node features is insufficient for obtaining compact representations, whereas our method is able to obtain highly compact but accurate representation which can be assured from the sufficiently high edge reconstruction accuracy. 
We believe that our proposed framework can not only learn accurate representations of nodes and edges, but also effectively compress their features, especially for large-scale real-world graphs, such as social networks or protein-protein interaction (PPI) graphs.

\subsection{Graph generation}
As shown in Figure~\ref{fig:concept} (left), graph generation depends heavily on the edge representations, as the model may generate incorrect graphs (e.g., toxic chemicals rather than drugs) if the edge information is inaccurate. Thus, we further validate our EHGNN on the graph generation tasks.

\paragraph{Experimental setup}
We directly forward the edge representation from the EHGNN to molecule generation networks, namely MolGAN~\cite{MolGAN} and MArkov moleculaR Sampling (MARS)~\cite{MARS}.
MolGAN uses the Generative Adversarial Network (GAN)~\cite{GAN}, to generate the molecular graph by balancing weights between its generator and discriminator.
MolGAN uses R-GCN~\cite{RelationalGCN} for node-level message-passing, whereas, for ours, we first obtain the edge representations using EHGNN, and use them with mean pooling in the graph encoder.
For evaluation metrics, we use the Synthetic Accessibility (SA) and Druglikeness (QED) scores.
We further apply EHGNN to MARS~\cite{MARS} that generates the molecule by sequentially adding or deleting its fragment, with MCMC sampling.
While the original model uses MPNN~\cite{MPNN} to implicitly obtain edge representations for adding and deleting actions, we use EHGNN to explicitly learn edge representations.
We train models to maximize four molecule properties: inhibition scores against two proteins, namely GSK3$\beta$ and JNK3 (biological); QED and SA scores (non-biological). Then we report the success rate at which the molecule satisfies all the properties. 
For more details, please see Appendix~\ref{sup/exp/gen}.

\paragraph{MolGAN results}
Figure~\ref{fig:molgan} shows the SA and QED scores of the generated molecules, of the MolGAN architecture with different encoders. Our EHGNN framework obtains significantly improved generation performance, over the original MolGAN which uses the R-GCN encoder and the MolGAN with GMPool, a state-of-the-art global node pooling encoder.
This is because EHGNN learns explicit edge representation which enhances the ability of the discriminator to distinguish between real and generated graphs.
The improvement in the discriminator also leads to notably more stable results compared to the baselines, which show a large variance in the quality of the generated molecules.

\paragraph{MARS results}
To perform correct editing actions to generate graphs with MARS, we need accurate edge representations, as edges determine the structure of the generated molecule. Table~\ref{tab:generation_MARS} shows that using our EHGNN achieves significantly higher generation performance over original MARS, which uses edges as auxiliary information only to enhance node representations. 
Notably, performance gain on the GSK3$\beta$ inhibition score for which structural binding is important, suggests that accurate learning of edges leads to generating more effective molecules that interact with the target protein.

\subsection{Graph and node classification}
Now, we validate the performance of our EHGNN with HyperDrop on classification tasks. Our approach is effective for the classification of graphs with or without edge features, since it allows lossless compression of nodes, and drops edges to allow message-passing only across relevant nodes.

\paragraph{Experimental setup}
Following the experimental setting of~\citet{gmt}, we use the GCN as the node-level message-passing layers for all models, and compare our edge pooling method against the existing graph pooling methods. 
For this experiment, our HyperDrop uses SAGPool~\cite{SAGPool} on the hypergraph, which is a node drop pooling method based on self-attention.
We use 6 datasets from the TU datasets~\cite{TUdataset/split} including three from the biochemical domains (i.e., DD, PROTEINS, MUTAG) and the remaining half from the social domains (i.e., IMDB-BINARY, IMDB-MULTI, COLLAB). Also, we further use the 4 molecule datasets (i.e., HIV, Tox21, ToxCast, BBBP) from the recently released OGB datasets~\cite{OGB}.
We evaluate the accuracy of each model with 10-fold cross validation~\cite{SortPool} on the TU datasets, and use ROC-AUC as the evaluation metric for the OGB datasets. For both datasets, we follow the standard experimental settings, from the feature extraction to the dataset splitting. We provide additional details of the experiments in Appendix~\ref{sup/exp/class}.

%%%%%%%%%%%%%%%%%%%%%%%%%%%%%%%%%%%%%%%%%%%%%%%%%%%%%%%%%%%%%
\begin{table*}[!t]
\caption{\small \textbf{Graph classification results}. The results are the mean and standard deviation over 10 different runs. Best performance and its comparable results ($p>0.05$) from the t-test are highlighted in bold. 
Hyphen (-) denotes out-of-resources that take more than 10 days. 
The results for the baselines are taken from \citet{gmt}.
}
\vspace{-0.075in}
\centering
\resizebox{\textwidth}{!}{
\renewcommand{\arraystretch}{1.0}
\renewcommand{\tabcolsep}{1.1mm}
\begin{tabular}{l l c c c c c c c c c c c}
\toprule
	& & \multicolumn{3}{c}{\textbf{TU : Biochemical}} & \multicolumn{3}{c}{\textbf{TU : Social}} & \multicolumn{4}{c}{\textbf{OGB : Molecule}} & \multirow{2}{*}{\textbf{Average}} \\
	\cmidrule(l{2pt}r{2pt}){3-5}
	\cmidrule(l{2pt}r{2pt}){6-8}
	\cmidrule(l{2pt}r{2pt}){9-12}
	 & & \textbf{D\&D} & \textbf{PROTEINS} & \textbf{MUTAG} & \textbf{IMDB-B} & \textbf{IMDB-M} & \textbf{COLLAB} & \textbf{HIV} & \textbf{Tox21} & \textbf{ToxCast} & \textbf{BBBP} \\
\midrule
	\# graphs & & 1178 & 1113 & 188 & 1000 & 1500 & 5000 & 41127 & 7831 & 8576 & 2039 \\
	\# classes & & 2 & 2 & 2 &  2 & 3 & 3 & 2 & 12 & 617 & 2 \\
	Avg \# nodes & & 284.32 & 39.06 & 17.93 & 19.77 & 13.00 & 74.49 & 25.51 & 18.57 & 18.78 & 24.06 \\  
	Avg \# edges & & 715.66 & 72.82 & 19.79 & 96.53 & 65.94 & 2457.78 & 27.47 & 19.27 & 19.26 & 25.95 \\
\midrule
\multirow{1}{*}{\textbf{Set}} & DeepSet & 77.39 $\pm$ 0.67 & 68.95 $\pm$ 0.92 & 72.56 $\pm$ 1.09 & 72.42 $\pm$ 0.36 & 50.24 $\pm$ 0.32 & 75.27 $\pm$ 0.21 & 71.20 $\pm$ 1.26 & 72.25 $\pm$ 0.23 & 59.44 $\pm$ 0.39 & 63.64 $\pm$ 0.62 & 68.34 \\
\cdashline{1-13}\noalign{\vskip 0.5ex}
\multirow{2}{*}{\textbf{Naive GNN}} & GCN & 72.05 $\pm$ 0.55 & 73.24 $\pm$ 0.73 & 69.50 $\pm$ 1.78 & 73.26 $\pm$ 0.46 & 50.39 $\pm$ 0.41 & 80.59 $\pm$ 0.27 & 76.81 $\pm$ 1.01 & 75.04 $\pm$ 0.80 & 60.63 $\pm$ 0.51 & 65.47 $\pm$ 1.73 & 69.70 \\ 
& GIN & 70.79 $\pm$ 1.17 & 71.46 $\pm$ 1.66 & 81.39 $\pm$ 1.53 & 72.78 $\pm$ 0.86 & 48.13 $\pm$ 1.36 & 78.19 $\pm$ 0.63 & 75.95 $\pm$ 1.35 & 73.27 $\pm$ 0.84 & 60.83 $\pm$ 0.46 & \textbf{67.65} $\pm$ 3.00 & 70.04 \\ 
\cdashline{1-13}\noalign{\vskip 0.5ex}
\multirow{2}{*}{\textbf{Global}} 
%& Set2Set & 71.94 $\pm$ 0.56 & 73.27 $\pm$ 0.85 & 69.89 $\pm$ 1.94 & 72.90 $\pm$ 0.75 & 50.19 $\pm$ 0.39 & 79.55 $\pm$ 0.39 & 74.70 $\pm$ 1.65 & 74.10 $\pm$ 1.13 & 59.70 $\pm$ 1.04 & 66.79 $\pm$ 1.05 & /10 \\ 
& SortPool & 75.58 $\pm$ 0.72 & 73.17 $\pm$ 0.88 & 71.94 $\pm$ 3.55 & 72.12 $\pm$ 1.12 & 48.18 $\pm$ 0.83 & 77.87 $\pm$ 0.47 & 71.82 $\pm$ 1.63 & 69.54 $\pm$ 0.75 & 58.69 $\pm$ 1.71 & 65.98 $\pm$ 1.70 & 68.49 \\
& GMT & \textbf{78.72} $\pm$ 0.59 & \textbf{75.09} $\pm$ 0.59 & 83.44 $\pm$ 1.33 & 73.48 $\pm$ 0.76 & 50.66 $\pm$ 0.82 & 80.74 $\pm$ 0.54 & \textbf{77.56} $\pm$ 1.25 & \textbf{77.30} $\pm$ 0.59 & \textbf{65.44} $\pm$ 0.58 & \textbf{68.31} $\pm$ 1.62 & 73.07 \\
\cdashline{1-13}\noalign{\vskip 0.5ex}
\multirow{7}{*}{\textbf{Hierarchical}} & DiffPool & 77.56 $\pm$ 0.41 & 73.03 $\pm$ 1.00 & 79.22 $\pm$ 1.02 & 73.14 $\pm$ 0.70 & \textbf{51.31} $\pm$ 0.72 & 78.68 $\pm$ 0.43 & 75.64 $\pm$ 1.86 & 74.88 $\pm$ 0.81 & 62.28 $\pm$ 0.56 & \textbf{68.25} $\pm$ 0.96 & 71.40 \\ 
& SAGPool & 74.72 $\pm$ 0.82 & 71.56 $\pm$ 1.49 & 73.67 $\pm$ 4.28 & 72.55 $\pm$ 1.28 & 50.23 $\pm$ 0.44 & 78.03 $\pm$ 0.31 & 71.44 $\pm$ 1.67 & 69.81 $\pm$ 1.75 & 58.91 $\pm$ 0.80 & 63.94 $\pm$ 2.59 & 68.49 \\
& TopKPool & 73.63 $\pm$ 0.55 & 70.48 $\pm$ 1.01 & 67.61 $\pm$ 3.36 & 71.58 $\pm$ 0.95 & 48.59 $\pm$ 0.72 & 77.58 $\pm$ 0.85 & 72.27 $\pm$ 0.91 & 69.39 $\pm$ 2.02 & 58.42 $\pm$ 0.91 & 65.19 $\pm$ 2.30 & 67.47 \\ 
& MinCutPool & \textbf{78.22} $\pm$ 0.54 & 74.72 $\pm$ 0.48 & 79.17 $\pm$ 1.64 & 72.65 $\pm$ 0.75 & 51.04 $\pm$ 0.70 & 80.87 $\pm$ 0.34 & 75.37 $\pm$ 2.05 & 75.11 $\pm$ 0.69 & 62.48 $\pm$ 1.33 & 65.97 $\pm$ 1.13 & 71.56 \\ 
& ASAP & 76.58 $\pm$ 1.04 & 73.92 $\pm$ 0.63 & 77.83 $\pm$ 1.49 & 72.81 $\pm$ 0.50 & 50.78 $\pm$ 0.75 & 78.64 $\pm$ 0.50 & 72.86 $\pm$ 1.40 & 72.24 $\pm$ 1.66 & 58.09 $\pm$ 1.62 & 63.50 $\pm$ 2.47 & 69.73 \\
& EdgePool & 75.85 $\pm$ 0.58 & \textbf{75.12} $\pm$ 0.76 & 74.17 $\pm$ 1.82 & 72.46 $\pm$ 0.74 & 50.79 $\pm$ 0.59 & - & 72.66 $\pm$ 1.70 & 73.77 $\pm$ 0.68 & 60.70 $\pm$ 0.92 & 67.18 $\pm$ 1.97 & - \\ 
& HaarPool & - & - & 66.11 $\pm$ 1.50 & 73.29 $\pm$ 0.34 & 49.98 $\pm$ 0.57 & - & - & - & - & 66.11 $\pm$ 0.82 & - \\ 
\midrule
\multirow{2}{*}{\textbf{Ours}} & HyperDrop & \textbf{78.74} $\pm$ 0.68 & \textbf{75.30} $\pm$ 0.45 & 84.00 $\pm$ 0.69 & 73.96 $\pm$ 0.41 & \textbf{51.68} $\pm$ 0.41 & \textbf{81.29} $\pm$ 0.25 & 76.79 $\pm$ 0.86 & 76.95 $\pm$ 0.32 & 64.21 $\pm$ 0.70 & \textbf{69.04} $\pm$ 0.86 & \textbf{73.20} \\ 
& HyperDrop + GMT & \textbf{78.39} $\pm$ 0.33 & \textbf{75.39} $\pm$ 0.26 & \textbf{85.72} $\pm$ 0.61 & \textbf{74.45} $\pm$ 0.61 & \textbf{51.45} $\pm$ 0.28 & 80.59 $\pm$ 0.33 & \textbf{77.84} $\pm$ 0.37 & \textbf{77.58} $\pm$ 0.43 & \textbf{65.15} $\pm$ 0.65 & \textbf{69.16} $\pm$ 1.04 & \textbf{73.57} \\
\bottomrule
\end{tabular}}
\label{tab:classification_hierarchical}
\vspace{-0.15in}
\end{table*}
%%%%%%%%%%%%%%%%%%%%%%%%%%%%%%%%%%%%%%%%%%%%%%%%%%%%%%%%%%%%%

\paragraph{Baselines}
We compare our EHGNN with HyperDrop, against the set encoding (DeepSet~\cite{deepsets}), GNNs with naive pooling baselines (GCN and GIN~\cite{GCN, GIN}), and state-of-the-art hierarchical pooling methods (DiffPool~\cite{DiffPool}, SAGPool~\cite{SAGPool}, TopKPool~\cite{TopKPool}, MinCutPool~\cite{MincutPool}, ASAP~\cite{ASAP}, EdgePool~\cite{edgepool}, and HaarPool~\cite{HaarPool}) that drop or coarsen node representations.
We also additionally compare or combine the state-of-the-art global node pooling methods (SortPool~\cite{SortPool}, GMT~\cite{gmt}) with our model, for example, HyperDrop + GMT. For more details, see Appendix~\ref{sup/exp/class}.

\begin{figure}[!t]
    \vspace{-0.05in}
    \begin{minipage}{0.33\linewidth}
        \centering
        \resizebox{\linewidth}{!}{
        \renewcommand{\arraystretch}{1.2}
        \renewcommand{\tabcolsep}{0.9mm}
        \begin{tabular}{l c c c}
        \toprule
            \textbf{Model} & \textbf{MUTAG} & \textbf{PROTEINS} & \textbf{Tox21} \\
        \midrule
            HyperDrop & 84.00 $\pm$ 0.69 & \textbf{75.30} $\pm$ 0.45 & \textbf{76.95} $\pm$ 0.32 \\
            HyperCluster & \textbf{84.50} $\pm$ 1.50 & 72.76 $\pm$ 1.12 & 76.68 $\pm$ 0.56 \\
        \midrule
            w/ RandDrop & 83.06 $\pm$ 1.15 & 74.92 $\pm$ 0.51 & 76.39 $\pm$ 0.47 \\
            w/o HyperDrop & 83.06 $\pm$ 1.20 & 75.08 $\pm$ 0.37 & 76.60 $\pm$ 0.45 \\
            w/o EHGNN & 69.50 $\pm$ 1.78 & 73.24 $\pm$ 0.73 & 75.04 $\pm$ 0.80 \\
        \bottomrule
        \end{tabular}}
        \vspace{-0.05in}
        \captionof{table}{\small \textbf{\small Ablation study} of HyperDrop on the MUTAG, PROTEINS, and Tox21 datasets for classification.
        }
        \label{tab:ablation}
    \end{minipage}
    \hfill
    \begin{minipage}{0.32\linewidth}
        \centering
        \centerline{\includegraphics[width=0.975\linewidth]{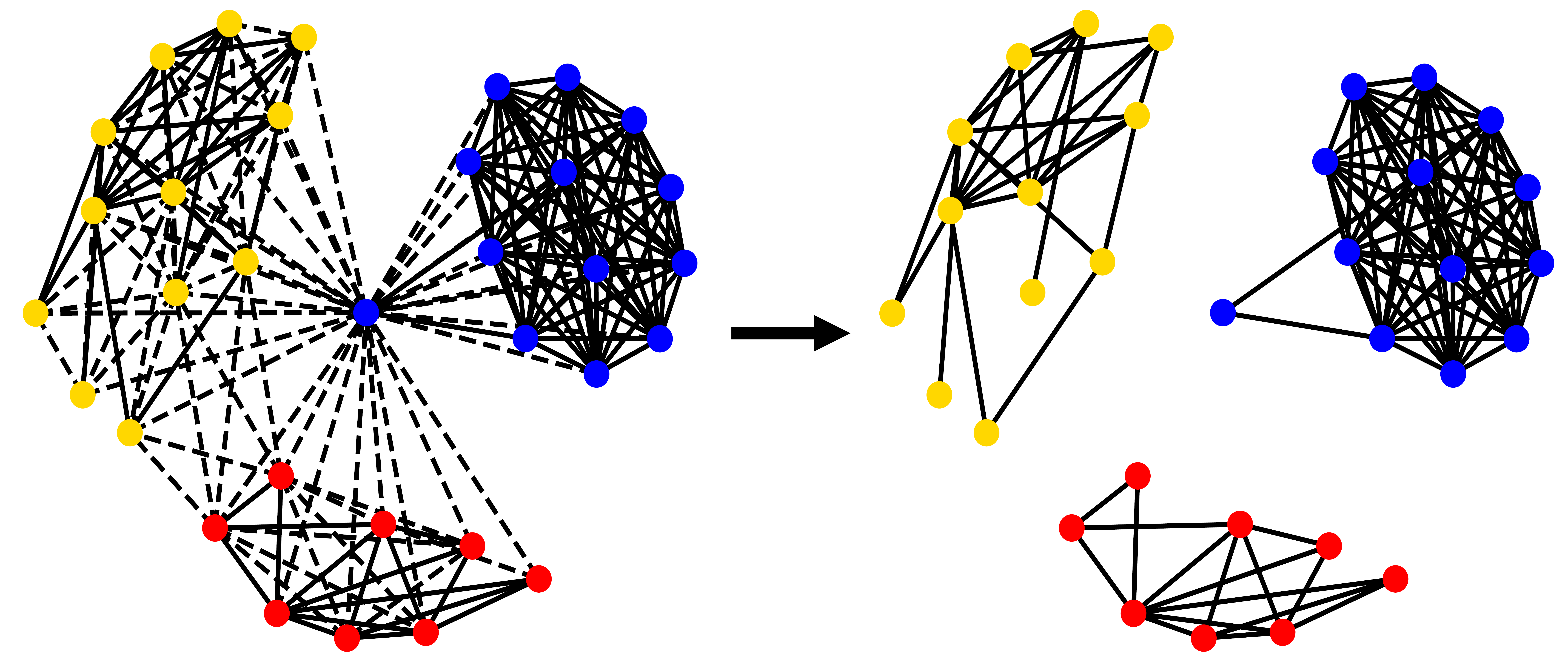}}
        \vspace{-0.05in}
        \caption{\small \textbf{\small Edge pooling results} on the COLLAB dataset. Colors denote connected components. 
        }
        \label{fig:drop}
    \end{minipage}
    \hfill
    \begin{minipage}{0.32\linewidth}
        \centering
        \includegraphics[width=1\linewidth]{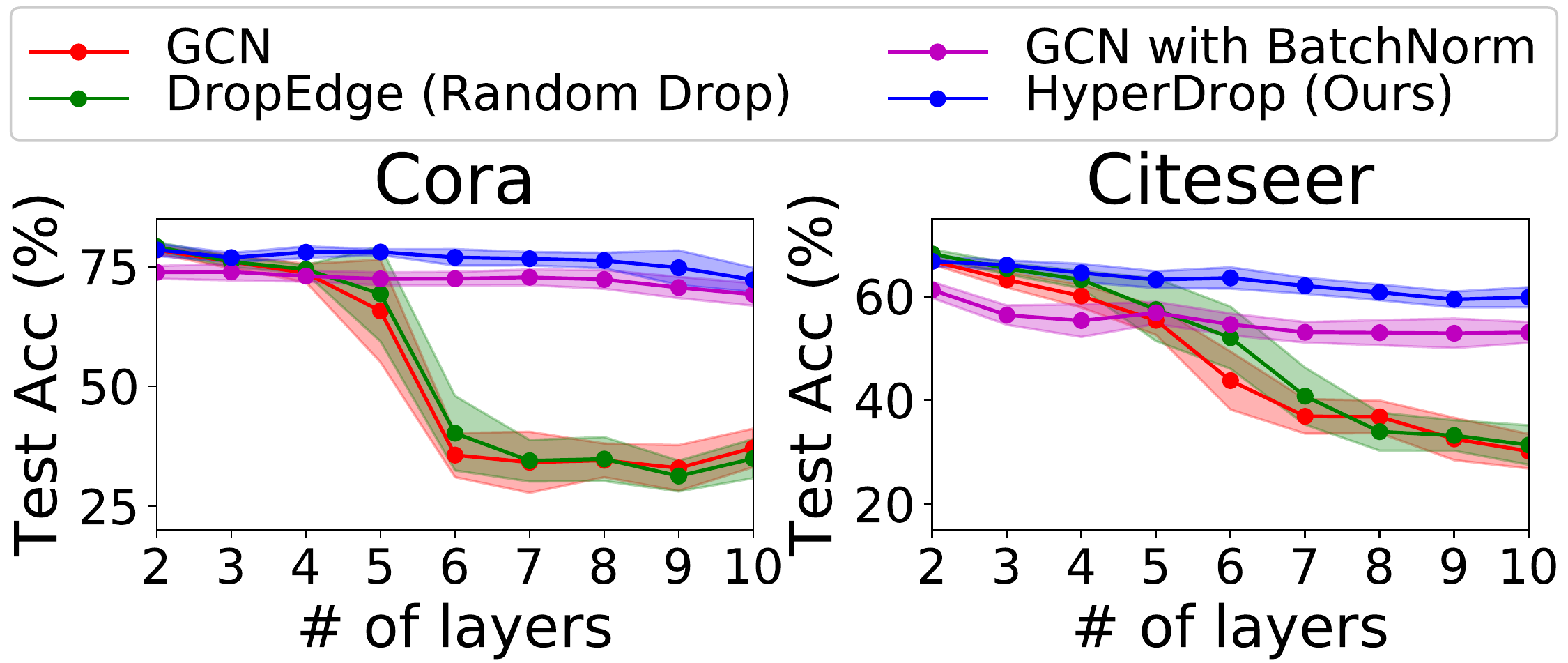}
        \vspace{-0.2in}
        \caption{\small \textbf{\small Node classification results.} Lines denote means over 10 runs and shades denote variances.
        }
    \label{fig:layers}
    \end{minipage}
    \vspace{-0.15in}
\end{figure}

\paragraph{Classification results}
Table~\ref{tab:classification_hierarchical} shows that the proposed EHGNN with HyperDrop significantly outperforms all hierarchical pooling baselines. This is because HyperDrop not only retains nodes by removing edges that are less useful for graph discrimination, but also explicitly uses the edge representations for graph classification. Since HyperDrop does not remove any nodes on the graph, it can be jointly used with any node pooling methods, and thus, we pair HyperDrop with GMT. This model largely outperforms GMT, obtaining the best performance on most of the datasets, which demonstrates that accurate learning of both the nodes and edges is important for classifying graphs.
We further visualize the edge pooling process of HyperDrop in Figure~\ref{fig:drop}, which shows that our method accurately captures the substructures of the entire graph, which leads to dividing the large graph into several connected components, thus adjusting the graph topology for more effective message-passing. We provide more visual examples of edge drop procedures in Appendix~\ref{sup/result/class}.

\paragraph{Ablation study}
To see how much each component contributes to the performance gain, we conduct an ablation study on EHGNN with HyperDrop. Table~\ref{tab:ablation} shows that, compared with a model that only uses node features (i.e., w/o EHGNN), learning explicit edge representations significantly improves performance. Our model EHGNN with HyperCluster, or without HyperDrop, or the model with random edge drop obtains decent performance, but substantially underperforms HyperDrop.  

%%%%%%%%%%%%%%%%%%%%%%%%%%%%%%%%%%%%%%%%%%%%%%%%%%%%%%%%%%%%%
%%%%%%%%%%%%%%%%%%%%%%%%%%%%%%%%%%%%%%%%%%%%%%%%%%%%%%%%%%%%%

\paragraph{Over-smoothing with deep GNNs}
Lastly, we demonstrate that our EHGNN with HyperDrop alleviates the over-smoothing problem of deep GNNs on semi-supervised node classification tasks, which is not possible for the existing node-based pooling methods. We follow the settings of existing works~\cite{GCN, GAT, grand} and provide the experimental details in Appendix~\ref{sup/exp/node}.
As shown in Figure~\ref{fig:layers}, HyperDrop retains the accuracy as the number of layers increases, whereas the naive GCN or random drop~\cite{DropEdge} results in largely degraded performance, since HyperDrop identifies and preserves the task-relevant edges while the sampling-based methods randomly drop the edges. 
Further, our method outperforms BatchNorm which alleviates over-smoothing by yielding differently normalized feature distribution at each batch. This is because HyperDrop splits the given graph into smaller subgraphs that capture meaningful message-passing substructures as shown in Figure~\ref{fig:drop}.

\section{Conclusion}
We tackled the problem of accurately representing the edges of a graph, which has been relatively overlooked over node representation learning. To this end, we proposed a novel edge representation learning framework using \emph{Dual Hypergraph Transformation} (DHT), which transforms the edges of the original graph into nodes on a hypergraph. This allows us to apply a message-passing scheme for node representation learning, for edge representation learning. Further, we proposed two edge pooling methods to obtain a holistic edge representation for a given graph, where one clusters similar edges into a single edge for graph reconstruction and the other drops unnecessary edges for graph classification. We validated our edge representation learning framework on graph reconstruction, generation, and classification tasks, showing its effectiveness over relevant baselines. 
\section{Acknowledgements and Disclosure of Funding}
We thank the anonymous reviewers for their constructive comments and suggestions. This work was supported by Institute of Information \& communications Technology Planning \& Evaluation (IITP) grant funded by the Korea government (MSIT) (No.2019-0-00075, Artificial Intelligence Graduate School Program (KAIST), and No.2021-0-02068, Artificial Intelligence Innovation Hub), and the Engineering Research Center Program through the National Research Foundation of Korea (NRF) funded by the Korean Government MSIT (NRF-2018R1A5A1059921).   

\bibliography{reference}

\newpage
\appendix
\vspace{0.3in}
\begin{center}{\bf {\LARGE Appendix \\
}}\end{center}
\vspace{0.4in}

\paragraph{Organization} 
The appendix is organized as follows. In section~\ref{sup/edge}, we first describe the structural details of the proposed \textit{Edge HyperGraph Neural Network} (EHGNN) framework using the \textit{Dual Hypergraph Transformation} (DHT) in comparison to those of the existing edge-aware graph neural networks. Also, we describe the detailed components of the proposed edge pooling methods: \textit{HyperCluster} and \textit{HyperDrop} in section~\ref{sup/pool}. We provide the experimental setups in Section~\ref{sup/exp}, which include the detailed descriptions of the models and datasets, as well as the experimental details for each task. Then, we provide additional experimental results on graph reconstruction and generation tasks with visualization of examples in Section~\ref{sup/result}. Finally, We discuss the limitations and potential societal impacts of our work in Section~\ref{sup/impact}.

\section{Edge Representation Learning \label{sup/edge}}
\vspace{-0.05in}
In this section, we first describe the detailed formulations of existing edge-aware graph neural networks (GNNs), and compare them with our methods. Then, we introduce the time complexity of making connectivity patterns for edges using existing edge-aware GNNs and our \textit{Edge HyperGraph Neural Network} (EHGNN). Finally, we discuss the sparse implementation of the proposed EHGNN along with our \textit{Dual Hypergraph Transformation} (DHT) at this end of the section.

\subsection{Discussion on edge-aware graph neural networks \label{sup/edge/aware}}
\vspace{-0.05in}
Here, we first formalize the existing edge-aware GNNs that we used as baselines~\cite{GCNOGB, MPNN, RelationalGCN, EdgeRepresentation/1}.
We begin by introducing the basic components of GNNs: $\bm{X}^{(l)}$ denotes the node features at $l$-th layer, $\bm{W}$ denotes the learnable weight matrix, $\bm{E}$ denotes the edge features, and $\mathcal{N}(v)$ denotes the neighboring node set of node $v$ in the given graph.

\vspace{-0.05in}
\paragraph{EGCN}
The node-wise formulation of edge-aware GCN~\cite{GCNOGB} is defined as follows:
\begin{equation}
    \bm{X}^{(l+1)}_v = \bm{W}\sum_{u\in\mathcal{N}(v)\cup\{v\}} n_{u,v}\cdot (\bm{X}^{(l)}_u + \bm{E}_{u,v})
\end{equation}
where $n_{u,v}$ is the normalizing coefficient for two adjacent nodes $u$ and $v$, and edge feature $\bm{E}$ is obtained from the categorical edge features without learning.

\vspace{-0.05in}
\paragraph{MPNN}
The node-wise formulation of MPNN~\cite{MPNN} using edge conditioned convolution~\cite{EdgeRepresentation/4} is defined as follows:
\begin{equation}
    \bm{X}^{(l+1)}_v = \bm{W}\bm{X}^{(l)}_v + \sum_{u\in\mathcal{N}(v)}\bm{X}^{(l)}_u\cdot\text{MLP}(\bm{E}_{u,v})
\end{equation}
where MLP is a linear layer for learning the edge representations to augment the node representations.

\vspace{-0.05in}
\paragraph{R-GCN}
The node-wise formulation of R-GCN~\cite{RelationalGCN} is defined as follows:
\begin{equation}
    \bm{X}^{(l+1)}_v = \bm{W}\bm{X}^{(l)}_v + \sum_{r\in\mathcal{R}}\sum_{u\in\mathcal{N}_r(v)} \frac{1}{|\mathcal{N}_r(v)|}\bm{W}_r\bm{X}^{(l)}_u
\end{equation}
where $\mathcal{R}$ is a set of categorical edge types,
%consisting of categorical values, 
and $\mathcal{N}_r(v)$ denotes the neighboring node set of the node $v$, having the associated edge type $r \in \mathcal{R}$.

\vspace{-0.05in}
\paragraph{EGNN}
The node-wise formulation of convolution-based edge GNN~\cite{EdgeRepresentation/1} is defined as follows:
\begin{equation}
    \bm{X}^{(l+1)}_v = \bm{W} \sum_{u\in\mathcal{N}(v)\cup\{v\}} \bm{E}^{(l)}_{u,v}\bm{X}^{(l)}_u
\end{equation}
where the edge features at $l$-th layer $\bm{E}^{(l)}$ are obtained by edge-level layers which are differently designed from node-level layers. The features are used as the attention coefficients for nodes to enhance the node-level representations. 

It is worthwhile to note that all baselines only implicitly capture edge information in the learned node representations rather than directly using it for downstream graph tasks, while our EHGNN framework explicitly learns and utilizes the learned edge representations.

\subsection{Sparse implementation of the dual hypergraph transformation \label{sup/edge/sparse}}
\vspace{-0.05in}
Since most graphs have relatively few connections per node, the number of non-zero elements in the adjacency matrix, which defines the connection among nodes, is smaller than the number of zero elements. Thus, using the adjacency matrix for message passing is highly inefficient in terms of memory usage.
To handle this issue, the most dominant approach is to use an edge list, which is a sparse representation of the adjacency matrix (or the incidence matrix) of the graph. Specifically, each column of the edge list $\bm{L}\in\mathbb{R}^{2\times m}$ denotes an edge $e$, which has two incident nodes $(v_{start}, v_{end})$, where $v_{start}$ denotes the start node and $v_{end}$ denotes the end node of the edge $e$.

Similarly, the incidence matrix of a hypergraph can be represented as a sparse form using a hyperedge list $\bm{L}^{\ast}\in\mathbb{R}^{2\times D}$, where $D$ is the sum of degrees of all nodes in the hypergraph. Each column of $L^{\ast}$ indicates a hyperedge $e^{\ast}$ with a (node, hyperedge) pair $(v^{\ast},e^{\ast})$, where $v^{\ast}$ is the node incident to the hyperedge $e^{\ast}$. If the hyperedge is incident to three nodes, then it will appear on three columns of $L^{\ast}$ paired with each incident node. Compared to this general hypergraph, the dual hypergraph obtained by DHT is 2-regular, which means each node in the hypergraph has a degree of two, since each edge in the original graph is incident to exactly two nodes.
Thanks to this property, the hyperedge list of the dual hypergraph has the dimensionality of $2\times 2m$ (i.e., $\bm{L}^{\ast}\in\mathbb{R}^{2 \times 2m}$). 

Then, the concrete implementation of DHT with the sparse edge list of the original graph and the sparse hyperedge list of its dual hypergraph is formalized as follows:
\begin{equation}
    \bm{DHT} \;:\; G = \big( \bm{X}, \bm{L}, \bm{E} \big) \;\mapsto\; G^{\ast} = \big( \bm{E}, \bm{L}^{\ast}, \bm{X} \big),
\end{equation}
where the hyperedge list $\bm{L}^{\ast}$ is obtained by reshaping the edge list $\bm{L}$ as follows:
\begin{align}
    &\bm{L}^{\ast}_{1,2i-1} = \bm{L}^{\ast}_{1,2i} = i, \\
    &\bm{L}^{\ast}_{2,2i-1} = \bm{L}_{1,i} \;,\;\; \bm{L}^{\ast}_{2,2i} = \bm{L}_{2,i},
\end{align}
for all $1\leq i\leq m$.

%%%%%%%%%%%%%%%%%%%%%%%%%%%%%%%%%%%%%%%%%%%%%%%%%%%%%%%%%%%
\begin{figure}[!t]
\begin{minipage}{0.58\linewidth}
\centering
\resizebox{\linewidth}{!}{
\renewcommand{\arraystretch}{1.3}
\renewcommand{\tabcolsep}{1.0mm}
    \begin{tabular}{lrrrrrr}
    \toprule
    \# of edges ($10^3$) & 2 & 4 & 8 & 16 & 32 & 64 \\
    \midrule
    Line graph & 32.78 & 65.93 & 131.36 & 260.92 & 527.44 & 1071.31 \\
    \midrule
    DHT (ours) & \textbf{0.13} & \textbf{0.18} & \textbf{0.26} & \textbf{0.45} & \textbf{0.81} & \textbf{1.37} \\
    \bottomrule
    \end{tabular}}
\captionof{table}{\small \textbf{Transformation time(s) results} of the line graph and our dual hypergraph. The results are the average of over 100 runs.}
\label{tab:transformation_complexity}
\end{minipage}
\hfill
\begin{minipage}{0.4\linewidth}
\centering
\resizebox{\linewidth}{!}{
\renewcommand{\arraystretch}{1.0}
\renewcommand{\tabcolsep}{0.8mm}
    \begin{tabular}{lcc}
    \toprule
    \multirow{2}{*}{Graph} & \multicolumn{2}{c}{Message-passing} \\
    & Nodes (GCN) & Edges (Ours) \\
    \midrule
    Erdos-Renyi & 0.0031 & 0.0034 \\
    \midrule
    Barabasi-Albert & 0.0029 & 0.0032 \\
    \bottomrule
    \end{tabular}}
\captionof{table}{\small \textbf{Message-passing time(s)} on the original graph and our dual hypergraph.
}
\label{tab:mp_complexity}
\end{minipage}
\vspace{-0.1in}
\end{figure}
%%%%%%%%%%%%%%%%%%%%%%%%%%%%%%%%%%%%%%%%%%%%%%%%%%%%%%%%%%%

\subsection{Complexity analysis \label{sup/edge/complexity}}
\vspace{-0.05in}
In this subsection, we provide the detailed complexity analysis of transformation and message-passing operations of existing edge-aware GNNs~\cite{EdgeRepresentation/0, EdgeRepresentation/1, EdgeRepresentation/2} and our DHT. We first introduce the transformation complexity, and then describe the message-passing complexity.

\vspace{-0.05in}
\paragraph{Transformation complexity}
To define the adjacency of edges to perform message-passing between edges, previous works either define the edge neighborhood structure~\cite{EdgeRepresentation/2}, or use the line graph transformation~\cite{EdgeRepresentation/0}. Constructing edge neighborhood takes $\mathcal{O}(m^2)$ for transforming the node adjacency to the edge adjacency, as, for verifying two edges are adjacent, we need to first sample one edge among $m$ edges, and then find the other edge that shares the same node among the remaining $m-1$ edges. In a similar manner, the complexity of line graph transformation is quadratic to the number of edges~\citet{linegraph/gcn}, as, for each pair of edges, we need to verify whether they share the same node. However, with our sparse implementation of DHT explained in~\ref{sup/edge/sparse}, we can obtain the hyperedge list -- a sparse data structure of the hypergraph -- by simply reshaping the given edge list of the original graph, which takes at most $\mathcal{O}(m)$.

We further experimentally verify the transformation complexity of the line graph transformation~\cite{EdgeRepresentation/0} and the proposed DHT, on Erdos-Renyi graph~\cite{randomgraph} with $1000$ nodes and the number of edges increasing from $2\times 10^3$ to $64\times 10^3$.
As shown in Table~\ref{tab:transformation_complexity}, our DHT is highly efficient compared to the line graph transformation, especially for large and dense graphs, as the line graph transformation is quadratic to the number of edges, whereas ours only requires simple tensor-reshape operations.

\vspace{-0.05in}
\paragraph{Message-passing complexity}
Note that the complexity of message-passing on the graphs depends only on the number of edges, thus it is enough to focus on the number of edges. 
When we transform the original graph into the line graph following~\citet{EdgeRepresentation/0}, the constructed line graph has $\mathcal{O}(m\cdot d_{max})$ edges, therefore the complexity of message-passing is $\mathcal{O}(m\cdot d_{max})$. For instance, when the input graph is a star graph having one hub node and $n$ other nodes (i.e., the number of edges is $n$), the line graph of the star graph has $n^2$ number of edges, thus the message-passing cost is $\mathcal{O}(n^2)$, as shown in Table 1 of the main paper. 
However, with our DHT implemented over the sparse hyperedge list, we only have $2m$ number of node-hyperedge pairs as explained in Section~\ref{sup/edge/sparse}, thus we can perform the message-passing between edges (nodes of the dual hypergraph) with complexity $\mathcal{O}(m)$. This complexity is equal to that of the message-passing between nodes of the original graph. In other words, the analytical complexity of message-passing between edges in equation 4 of the main paper is equivalent to the complexity of message-passing between nodes in equation 1 of the main paper.

We experimentally validate the message-passing complexity on the original graph (message-passing between nodes) and the dual hypergraph (message-passing between edges) in Table~\ref{tab:mp_complexity}. 
We evaluate the message-passing time on both the Erdos-Renyi graph~\cite{randomgraph} and the scale-free (Barabasi-Albert) network~\cite{scalefree}, with $3000$ nodes $11984$ edges following the densification law (i.e. $m\propto n^{1.18}$~\cite{densification}) of the internet graph.
Table~\ref{tab:mp_complexity} shows that message-passing time on the dual hypergraph is almost equal to the message-passing time on the original graph, which coincides with the previous analysis.

\section{Details for Edge Pooling Schemes \label{sup/pool}}
\vspace{-0.05in}
In this section, we describe the proposed two novel edge pooling schemes: \textit{HyperCluster} that coarsens similar edges for global edge representations, and \textit{HyperDrop} that drops unnecessary edges for hierarchical graph representations.

\subsection{HyperCluster \label{sup/pool/cluster}}
Our cluster-based edge pooling model, HyperCluster, consists of edge-level message passing layers (i.e., EHGNN layers) and HyperCluster layers, which we describe below in detail. Before clustering edges, we first update the edge features using multiple EHGNN layers as follows:
\begin{equation}
    \bm{E}^{(l+1)}=\text{EHGNN}(\bm{X},\bm{M},\bm{E}^{(l)}),
\end{equation}
where $\bm{E}^{(l)}$ denotes the updated edge features at the $l$-th layer from the initial edge features $\bm{E}^{(0)}=\bm{E}$, and we finally obtain $\bm{E'} = \bm{E}^{(L)}$ after $L$ number of EHGNN layers.
Then, to obtain the global edge representation of the entire graph, we cluster the nodes of its dual hypergraph using the node clustering method. 
While we can use any off-the-shelf node clustering methods~\cite{DiffPool, MincutPool, gmt}, in this paper, we use the state-of-the-art pooling method, namely GMPool~\cite{gmt}. To apply GMPool on a hypergraph, we modify the graph multi-head attention block (GMH), which is used to construct key and value matrices using GNNs for the original graph structure in the GMPool paper~\cite{gmt}, for the hypergraph structure by replacing the adjacency matrix to the incidence matrix.
We compress $m$ nodes in the dual hypergraph into $k$ nodes with the modified $\text{GMPool}_k$, formalized as follows:
\begin{equation}
    \bm{E}^{pool} = \text{GMPool}_k(\bm{E'},\bm{M}^T), \;\; \bm{M}^{pool} = \bm{M}\bm{C},
\end{equation}
where $\bm{C}$ is the cluster assignment matrix generated by GMPool.
The overall architecture can be either global or hierarchical, depending on the downstream task.

\subsection{HyperDrop \label{sup/pool/drop}}
Our drop-based edge pooling model, HyperDrop, consists of EHGNN layers and HyperDrop layers, which we describe below in detail.
Before dropping unnecessary edges, we first update the edge features using the proposed EHGNN layer as follows:
\begin{equation}
    \bm{E'}=\text{EHGNN}(\bm{X},\bm{M},\bm{E}).
\end{equation}
Then, we drop the nodes of the dual hypergraph based on a learnable score function. While we can use any off-the-shelf node drop methods~\cite{TopKPool, SAGPool} with their score functions, in this paper, we use the self-attention score based node drop method proposed in \citet{SAGPool} as follows:
\begin{equation}
    \bm{Z} = \text{tanh}(\text{GNN}(\bm{E'}, \bm{M}^T, \bm{X}))
\end{equation}
Based on the output score vector $\bm{Z} \in \mathbb{R}^{m}$ for every $m$ nodes on the dual hypergraph, we select the top-ranked $k$ nodes to obtain the pooled edge features and the incidence matrix as follows: 
\begin{equation}
    \bm{E}^{pool} =  \bm{E'}_{idx} \,, \;\;
    \bm{M}^{pool} = ((\bm{M}^T)_{idx})^T \;;\;
    idx = \text{top}_k\left(\bm{Z}\right).
\end{equation}
Thus, we obtain the edge-pooled graph $G^{pool} = (\bm{X},\bm{M}^{pool},\bm{E}^{pool})$ without loss of node information of the original graph.
Furthermore, we use the self-attention score vector $\bm{Z}$ as the edge weight for the node-level message passing layer, to reflect the relative importance of the neighboring information. This can be formulated as follows:
\begin{equation}
    \bm{X'} = \text{GNN}\left( \bm{X}, \bm{M}^{pool}, \bm{Z}_{idx} \right),
\end{equation}
where we can use simple GCN~\cite{GCN} or edge-aware GNNs for the GNN function.

\section{Experimental Setup \label{sup/exp}}
\vspace{-0.05in}
In this section, we introduce baselines and proposed models that we used for verifying the effectiveness of our approaches, in two different paragraphs: one for message passing methods and another for graph pooling methods, and then provide the information of the computing resources. After that, we describe the experimental details about four different tasks on which we validate our methods.

\vspace{-0.05in}
\paragraph{Baselines and our model for graph neural networks \label{sup/exp/baseline}}
Here, we describe a set encoding model that ignores connectivity between nodes, naive graph neural networks that only consider node features without edge information, edge-aware graph neural networks that use edge features as auxiliary information for updating node features, and our model that explicitly represents edges as follows:

\vspace{-0.05in}
\begin{enumerate}[itemsep=1.0mm, parsep=1pt, leftmargin=*]
    \item {\bf DeepSet.} This method~\cite{deepsets} is the set encoding baseline that first represents each node with a linear function, and then aggregates all node representations with sum pooling, which does not consider connectivity patterns between nodes.
    
    \item {\bf GCN.} This method~\cite{GCN} is the naive graph neural network baseline that aggregates neighboring nodes' information using the mean operation, which does not consider edge information. Also, we obtain the entire graph representation using the mean pooling of all nodes.
    
    \item {\bf GIN.} This method~\cite{GIN} is the naive graph neural network baseline that aggregates neighboring node's information using the sum operation, which does not consider edge information. Also, we obtain the entire graph representation using the sum pooling of all nodes.
    
    \item {\bf EGCN.} This method~\cite{OGB} is the edge-aware graph neural network baseline that uses edges as auxiliary information only to augment node-level representations, by adding the edge features between a node and its neighborhood to the node features (see Section~\ref{sup/edge/aware} for detailed formulation).
    
    \item {\bf MPNN.} This method~\cite{MPNN} is the edge-aware graph neural network baseline that uses edges as auxiliary information only to augment the node-level representations, by multiplying the edge features between a node and its neighborhood to the node feature (see Section~\ref{sup/edge/aware} for details).
    
    \item {\bf R-GCN.} This method~\cite{RelationalGCN} is the edge-aware graph neural network baseline that uses discrete edge features for considering relation types between nodes, by multiplying the categorical weights of edges to the node features (see Section~\ref{sup/edge/aware} for detailed formulation).
    
    \item {\bf EGNN.} This method~\cite{EdgeRepresentation/1} is the edge-aware graph neural network baseline that first obtains explicit edge representations using differently designed edge-level layer, and then uses them to augment node-level representations, by multiplying the edge representations to the node representations (see Section~\ref{sup/edge/aware} for detailed formulation).
    
    \item {\bf EHGNN.} This is our edge representation learning framework that first transforms the given original graph into its dual hypergraph with \textit{Dual Hypergraph Transformation}, and then obtain the explicit edge representations with existing off-the-shelf message-passing schemes for nodes, which is further directly used for graph-level representation learning. 
\end{enumerate}

\vspace{-0.05in}
\paragraph{Baselines and our model for graph pooling}
Here, we explain the global node pooling baselines, as well as the hierarchical node pooling baselines. Then, we describe the proposed two novel edge pooling schemes: cluster-based and drop-based methods, for graph-level representation learning.

\vspace{-0.05in}
\begin{enumerate}[itemsep=1.0mm, parsep=1pt, leftmargin=*]
    \item {\bf DiffPool.} This method~\cite{DiffPool} is the hierarchical node pooling baseline that coarsens nodes with a clustering-based approach, where it generates a cluster-assignment matrix for nodes using a GNN.
    
    \item {\bf SAGPool.} This method~\cite{SAGPool} is the hierarchical node pooling baseline that drops unnecessary nodes with a drop-based approach, where it generates scores for nodes with a GNN.
    
    \item {\bf TopKPool.} This method~\cite{TopKPool} is the hierarchical node pooling baseline that drops unnecessary nodes with a drop-based approach, where it generates scores for nodes with MLPs.
     
    \item {\bf MinCutPool.} This method~\cite{MincutPool} is the hierarchical node pooling baseline that coarsens nodes with a clustering-based approach, where it generates a cluster-assignment matrix for nodes using MLPs.
    
    \item {\bf ASAP.} This method~\cite{ASAP} is the hierarchical node pooling baseline that first clusters similar nodes, then drop unnecessary clusters to coarsen an entire graph.
    
    \item {\bf EdgePool.} This method~\cite{edgepool} is the hierarchical node pooling baseline that computes the edge score between nodes, then contracts two adjacent nodes with the high edge score into a single node.
    
    \item {\bf HaarPool.} This method~\cite{HaarPool} is the hierarchical node pooling baseline that coarsens nodes with the Haar transformation, which is based on the Haar basis in the Haar wavelet domain~\cite{haar}.
    
    \item {\bf SortPool.} This method~\cite{SortPool} is the global node pooling baseline that first sorts the obtained node representations at the end of graph convolution layers, then predicts an entire graph representation with sorted node features.
    
    \item {\bf GMPool.} This method~\cite{gmt} is the global node pooling baseline that uses self-attention based operations to compress multiple nodes into a few clusters with learnable cluster assignment vectors to obtain an entire graph representation.
    
    \item {\bf GMT.} This method~\cite{gmt} is the global node pooling baseline that stacks self-attention based layers not only to compress many nodes into a few clusters with learnable cluster assignment vectors, but also to consider the inter-node (or cluster) relationships to obtain an entire graph representation.
    
    \item {\bf HyperCluster.} This is our global edge representation learning scheme that coarsens similar edges into a single edge to obtain a holistic edge-level representation, where we can generate the cluster assignment matrix for edges using existing clustering-based methods, such as GMPool~\cite{gmt} (see Section~\ref{sup/pool/cluster} for more details).
    
    \item {\bf HyperDrop.} This is our hierarchical edge representation learning scheme that drops unnecessary edges based on a learnable score function, such as MLPs or GNNs, thereby adjusting the graph topology for more effective message passing. Notably, this scheme does not result in the removal of any nodes.
    (see Section~\ref{sup/pool/drop} for more details).
\end{enumerate}

\vspace{-0.05in}
\paragraph{Computing resources}
For all experiments, we use PyTorch~\cite{pytorch} and PyTorch geometric~\cite{pytorchgeo}, and train each model on a single Titan XP, GeForce GTX Titan X, or GeForce RTX 2080 Ti GPU. A single experiment of each task takes less than 1 day, and for the classification tasks such as node or graph classification, the single runtime on most datasets of a relatively small size is less than 1 hour.

\subsection{Graph reconstruction \label{sup/exp/recon}}
\vspace{-0.05in}

\paragraph{Common implementation details}
Given a set of graphs $\left\{G=(\bm{X},\bm{M},\bm{E})\right\}$, the goal of graph reconstruction is to reconstruct both node and edge features from the compressed representations, by training two separate autoencoders where one is trained for reconstructing node features and the other is trained for reconstructing edge features. Formally, we define the node and edge encoders as $\text{ENC}_{node}$ and $\text{ENC}_{edge}$, respectively, and the node and edge decoders as $\text{DEC}_{node}$ and $\text{DEC}_{edge}$, respectively. Then, following the standard architecture setting of graph reconstruction tasks of existing works~\cite{MincutPool, gmt}, the node-level autoencoder which is a pair of the node encoder and node decoder, $\text{ENC}_{node}$ and $\text{DEC}_{node}$, is defined as follows:
\begin{align}
    & \text{ENC}_{node}(\bm{X}, \bm{M}, \bm{E}) 
    = \text{GMPool}( \text{GNN}( \text{GNN}( \bm{X}, \bm{M}, \bm{E} ) ) ) 
    = \bm{X}^{pool}, \\
    & \text{DEC}_{node}(\bm{X}^{pool}, \bm{M}, \bm{E})
    = \text{GNN}(\text{GNN}(\text{GNN}( \text{GMPool}^{-1}(\bm{X}^{pool}, \bm{M}, \bm{E}) )))
    = \bm{X}^{rec},
\end{align}
where we use the GMPool~\cite{gmt} for reconstructing node features, as it shows outstanding performance on node-level reconstruction tasks. $\text{GMPool}$ denotes the pooling operation, and $\text{GMPool}^{-1}$ denotes the unpooling operation following the setting of the original paper~\cite{gmt}. Also, $\bm{X}^{rec} \in \mathbb{R}^{n\times d}$ is the reconstructed node features from the pooled node representations $\bm{X}^{pool} \in \mathbb{R}^{k\times d}$, where $k$ is the number of pooled nodes and $n$ is the number of all nodes. We omit the inputs of the GNN, which are the incidence matrix $\bm{M}$ and the edge feature matrix $\bm{E}$, for simplicity.

However, to reconstruct the entire graph which have both node and edge features, we further need to define a separate edge-level autoencoder. Thus, similarly to the node-level autoencoder, we define the edge-level reconstruction module as a pair of the edge encoder and edge decoder, $\text{ENC}_{edge}$ and $\text{DEC}_{edge}$, formalized as follows:
\begin{align}
    & \text{ENC}_{edge}(\bm{X},\bm{M}, \bm{E}) 
    = \text{Pool}( \text{GNN}( \text{GNN}( \bm{X},\bm{M}, \bm{E} ) ) ) 
    = \bm{E}^{pool}, \\
    & \text{DEC}_{edge}(\bm{X}, \bm{M}, \bm{E}^{pool})
    = \text{GNN}(\text{GNN}(\text{GNN}( \text{Pool}^{-1}(\bm{X}, \bm{M}, \bm{E}^{pool}) )))
    = \bm{E}^{rec},
\end{align}
where, for our models, we use the EHGNN with the GCN~\cite{GCN} for GNN operations, and HyperCluster for pooling and unpooling operations which is described in Section~\ref{sup/pool/cluster} in detail.
Meanwhile, for the baselines, we use the existing edge-aware GNNs~\cite{GCNOGB, MPNN, RelationalGCN, EdgeRepresentation/1} for GNN operations, and GMPool~\cite{gmt} for pooling and unpooling operations, where we obtain the final edge representation by averaging the two representations of incident nodes for the edge. 
This is because the baselines only use edge features as auxiliary information for updating node features. $\bm{E}^{rec} \in \mathbb{R}^{m\times d'}$ is the reconstructed edge features from the pooled edge representations $\bm{E}^{pool} \in \mathbb{R}^{k'\times d'}$, where $k'$ is the number of pooled edges and $m$ is the number of all edges. Similar to the formulation of the node-level autoencoder, we omit the inputs of the GNN for simplicity.

Our reconstruction objective is to minimize the discrepancy between the original graph $G=(\bm{X},\bm{M},\bm{E})$ and the reconstructed graph $G^{rec}=(\bm{X}^{rec},\bm{M},\bm{E}^{rec})$, with a loss function such as mean squared error or cross-entropy loss for node and edge features.
For the edge reconstruction task, we only use the edge autoencoder without using the node autoencoder. For all reconstruction experiments, the learning rate of the node autoencoder is set to $5\times 10^{-3}$, and the learning rate of the edge autoencoder is set to $1\times 10^{-3}$. We optimize the full network using an Adam optimizer~\cite{kingma2014adam}.

\vspace{-0.05in}
\paragraph{Implementation details on synthetic graphs}
For the edge reconstruction of a synthetic graph, we use the standard two-moon graph generated by the PyGSP library~\cite{pygsp}, with node features given by their coordinates and edge features given by RGB colors of which values range from 0 to 1. Then, the goal of the edge reconstruction task is to restore all edge colors from the compressed edge representations after edge pooling. To minimize the discrepancy between original and reconstructed edge features, we use the mean squared error loss as the learning objective. Also, we use the early stopping criterion, where we stop the training if there is no further improvement on the training loss during 1,000 epochs, and the maximum number of epochs is set to 5,000. We set the pooling ratio of all models as 1\% with the hidden dimension of size 16.

\vspace{-0.05in}
\paragraph{Implementation details on molecular graphs}
Following the experimental setting of the existing work~\cite{benchmarkingGNN, gmt}, we use the subset of the full ZINC dataset~\cite{ZINC}, which consists of 12K molecular graphs, where node features are atom types and edge features are bond types. The number of atom types is 28, and the number of bond types is 5. We follow the dataset splitting of training, validation, and test sets from~\citet{benchmarkingGNN}. Then, the goal of the molecular graph reconstruction task is to restore both atom types and bond types of all nodes and edges from their compressed representations after pooling. To train the model, we use the cross-entropy loss for molecular graph reconstruction, since the initial features given for nodes and edges are discrete. We also use the early stopping criterion, where we stop the training if there is no further improvement on the validation loss during 200 epochs. For hyperparameters, the maximum number of epochs is set to 500, hidden dimension size is set to 32, and batch size is set to 128. We run five experiments with different random seeds, and report the average performance with its standard deviation. Following the evaluation setup of~\citet{gmt}, we use the following three metrics: \emph{accuracy} measures the classification accuracy of all nodes and edges, \emph{validity} counts the number of reconstructed molecules which are chemically valid, and \emph{exact match} counts the number of reconstructed molecules which are identical to the original molecules.

\vspace{-0.05in}
\paragraph{Implementation details on graph compression}
We quantitatively compare the relative memory size of the compressed graph after pooling nodes and edges against the size of the original graph, which we use the Erdos-Renyi random graph model~\cite{randomgraph}. We compare our proposed method EHGNN with HyperCluster, with the node pooling baseline, GMT.
The number of nodes is fixed to $10^3$, while the number of edges is selected from one of $10^3$, $5\times 10^3$, and $10^4$. To obtain the features of nodes and edges, we first randomly assign one of three values to each node (i.e., one among $\left\{ 0, 1, 2 \right\}$), and then generate edge features using the values of two adjacent nodes for each edge. For example, if two nodes have the same 0 value for the incident edge, then we assign the zero value to the edge feature. Since the total number of pairs of node values is six for the undirected graph, the number of edge features is six. 
The node pooling ratio is equally fixed to 15\% for both GMT and our model, and we report the relative size of the entire graph with the edge reconstruction accuracy higher than 95\% or 75\%, where the edge pooling ratio is decided according to its accuracy.

\subsection{Graph generation \label{sup/exp/gen}}
\vspace{-0.05in}

\paragraph{Implementation details on MolGAN architectures}
We use the QM9 dataset~\cite{ramakrishnan2014quantum} that contains 133,885 organic compounds, where each molecular graph consists of carbon (C), oxygen (O), nitrogen (N), and fluorine (F) with up to nine non-hydrogen atoms. To evaluate the generated molecular graphs, we use the normalized Synthetic Accessibility (SA) and Druglikeness (QED) scores following the evaluation setup of the original paper~\cite{MolGAN}. Also, we use the categorical re-parameterization trick with the Gumbel-softmax function during the discretization process of molecule generation, to train the model in an end-to-end fashion, which adapts the learning scheme of the original paper~\cite{MolGAN}.

In the original MolGAN~\cite{MolGAN}, R-GCNs~\cite{RelationalGCN} are used to encode feature representations of nodes for the discriminator and reward networks. Learning rates of the generator, the discriminator, and the reward network are equally set to $1\times 10^{-3}$, and hidden sizes of the two-layer R-GCNs are 128 and 64.
For the MolGAN with GMPool (MolGAN + GMPool) setting, the GMPool, which is the global node pooling baseline, is additionally used to obtain the compact node-level representations. The $\text{tanh}$ activation function is used for GMPool.
For the MolGAN with the proposed EHGNN (MolGAN + EHGNN) setting, we use two EHGNN layers to encode the feature representations of the edges, wherein we use the GCN as the edge-level message-passing function. The hidden sizes are set to 32 and 16. After obtaining the edge-level representations, we use mean pooling to obtain the global edge representation, which is forwarded to the discriminator and reward networks.
We further combine the GMPool with the MolGAN + EHGNN combination (MolGAN + GMPool + EHGNN) to additionally enhance the global graph representation with both node and edge representations.
The learning rate of the EHGNN parameters in the discriminator and reward networks is set to $1\times 10^{-2}$.
Also, all the models use Adam optimizer~\cite{kingma2014adam} for training. Regarding other settings, we strictly follow the original MolGAN paper~\cite{MolGAN}, and use the available code\footnote{https://github.com/yongqyu/MolGAN-pytorch}.

%%%%%%%%%%%%%%%%%%%%%%%%%%%%%%%%%%%%%%%%%%%%%%%%%%%%%%%%%%%%%%%%%%%%%%%%%%%%%%%%%
\begin{wraptable}{t}{0.33\textwidth}
    \vspace{-0.16in}
    \centering
    \captionof{table}{\small Statistics of fragment vocabularies of ZINC15 and ChEMBL datasets on MARS experiments.}
    \vspace{-0.1in}
    \resizebox{0.33\textwidth}{!}{
        \renewcommand{\tabcolsep}{0.75mm}
        \renewcommand{\arraystretch}{0.8}
        \begin{tabular}{lcc}
        \toprule
             & \textbf{ZINC15} & \textbf{ChEMBL} \\
        \midrule
            \# of node types & 9 & 9 \\
            Avg \# of nodes & 7.68 & 7.35 \\
            \# of edge types & 4 & 4 \\
            Avg \# of edges & 7.54 & 7.08 \\
        \bottomrule
        \end{tabular}
    }
    \label{tab:MARS_dataset}
    \vspace{-0.15in}
\end{wraptable}
%%%%%%%%%%%%%%%%%%%%%%%%%%%%%%%%%%%%%%%%%%%%%%%%%%%%%%%%%%%%%%%%%%%%%%%%%%%%%%%%%

\vspace{-0.05in}
\paragraph{Implementation details on MARS architectures}
For the experiments using the MARS architecture, we use the ZINC15~\cite{ZINC15, pretrain-gnns} dataset, which contains 2 million molecules, and we use the available data\footnote{http://snap.stanford.edu/gnn-pretrain/data/} from~\citet{pretrain-gnns}.
Further, we provide additional experimental results on the ChEMBL~\cite{gaulton2017chembl} dataset, which consists of 1,488,640 molecules, in Section~\ref{sup/result/gen}.
As the fragments of molecular graphs are the basic building blocks for molecular graph generation in the MARS~\cite{MARS}, we build the fragment vocabularies following the same procedure of the original MARS paper: fragments are built by breaking a single bond of molecules from the given dataset, limiting the size of fragments to 10 atoms (see the original paper~\cite{MARS} for more details on the generation process of fragment vocabularies). We report the statistics of generated fragments from each dataset in Table~\ref{tab:MARS_dataset}.

The MARS model sequentially generates molecules by taking one of the addition or deletion actions at each step, especially where this model uses the explicit edge representation on the deletion actions. For a set of given graphs $\{G=(\bm{X},\bm{M},\bm{E})\}$, the original MARS model obtains the edge representation for the deletion actions as follows:
\begin{align}
\begin{split}
    \bm{X'} &= \text{MPNN}(\bm{X}, \bm{M}, \bm{E}) \\
    \bm{E'}_{e} &= \text{Concat}(\bm{X'}_{u}, \bm{X'}_{v}, \text{MLP}(\bm{E}_e))
\end{split}
\end{align}
where MPNN is the edge-aware graph neural network described in the subsection~\ref{sup/exp/baseline}, an edge $e$ is incident to two nodes $u$ and $v$, and $\bm{E'}_{e}$ is the output edge representation of the edge $e$. Compared to this baseline that implicitly captures the edge representation on the learned node representation $\bm{X'}$ with the concatenated edge representation through the naive MLP layer, for our model, we replace the MLP layer with the proposed EHGNN to explicitly learn the edge representation via edge-level message passing.
For a fair comparison in terms of the number of parameters, we use the same number of layers and embedding size for both MLP and EHGNN.

Following the experimental setup of the original MARS paper~\cite{MARS}, we train the models to maximize the sum of multiple scores: QED, SA, and target protein inhibition scores against GSK3$\beta$ and JNK3, respectively. For evaluation metrics, we measure the percentage of the generated molecules having scores above a certain threshold for each property: QED $\ge 0.67$, SA $\ge 0.67$, and the inhibition scores against GSK3$\beta$ $\ge 0.6$ and JNK3 $\ge 0.6$. The success rate can measure the overall multi-objective score by calculating the percentage of the generated molecules satisfying all four objectives. We also report the suggested easier threshold from the original MARS paper~\cite{MARS}: QED $\ge 0.6$, SA $\ge 0.67$, and the inhibition scores against GSK3$\beta$ $\ge 0.5$ and JNK3 $\ge 0.5$, in Section~\ref{sup/result/gen}, where we see the same tendency for the results of baseline and our model. For the experiment on ZINC15, we set the learning rate of EHGNN parameters to $5 \times 10^{-3}$ with a cosine scheduler for learning rate warmup. For the experiment on ChEMBL, we set the learning rate of EHGNN parameters to $3 \times 10^{-4}$. The learning rate of other parameters in MARS is set to $3 \times 10^{-4}$, following the original paper~\cite{MARS}. We use the available code\footnotemark[\getrefnumber{footnote:mars}] from the original MARS paper.

\subsection{Graph classification \label{sup/exp/class}}
\vspace{-0.05in}

\paragraph{Datasets}
We validate our models on ten different benchmark datasets including six from the TU datasets~\cite{classification/datasets} and four from the OGB datasets~\cite{OGB}.
For a fair comparison of baselines and our model, following the standard experimental setting of~\citet{fair/GNN}, we use the one-hot encoding of atom types as initial node features in TU bio-chemical datasets (D\&D, PROTEINS, MUTAG) and one-hot encoding of node degrees as initial node features in TU social datasets (IMDB-B, IMDB-M, COLLAB), if initial node features are not given in advance. Furthermore, if the initial edge features are not given in advance, we set them to one uniformly.
For the dataset splitting of the TU datasets, we follow the standard training/test splits from~\citet{TUdataset/split, SortPool, gmt}, and further divide the training set into training and validation sets by using the 10 percent of the training data as validation data, as suggested by the fair comparison setup of~\citet{fair/GNN}.
For the OGB datasets (HIV, Tox21, ToxCast), following the original dataset paper~\cite{GCNOGB}, we use the additional atom and bond features for each graph, and follow the performance evaluation and data split setting of \citet{GCNOGB}. 
The statistics of each dataset are provided in Table 3 of the main paper.

\vspace{-0.05in}
\paragraph{Implementation details}
We follow the standard experiment setting from~\citet{gmt} with the same base architectures and hyperparameters for all models on all datasets\footnote{https://github.com/JinheonBaek/GMT}. Notably, we stack three number of GCN layers as node-level message passing for all pooling models, including ours. 
For our model, we use the GCN for the EHGNN layer, where we equally stack three number of EHGNN layers to obtain the explicit edge representations, in parallel with node-level layers. Also, from the explicitly learned edge representations, we drop edges with their scores at each edge-level layer, which is described in section~\ref{sup/pool/drop} in detail.
For the model HyperDrop + GMT, we apply the global node pooling layer GMPool~\cite{gmt} after the HyperDrop layers to obtain the global representation.
For the hyperparameters of our HyperDrop, we set the hidden dimension of edges as 128 except the COLLAB dataset, on which we set the hidden dimension as 16, since the COLLAB dataset has a large number of edges compared to other datasets. Also, we randomly search for the edge drop ratio by increasing the drop ratios from $5\%$ to $75\%$ with $5\%$ increments. We report the average performances and standard deviations of 10 runs with different random seeds on test datasets.

\subsection{Node classification \label{sup/exp/node}}
\vspace{-0.05in}
To demonstrate HyperDrop's effectiveness in alleviating the over-smoothing problem in deep GNNs, we validate it on the semi-supervised node classification tasks. 

\vspace{-0.05in}
\paragraph{Datasets} We experiment on two benchmark datasets~\cite{datasets/node}, namely Cora and Citeseer, which is the citation network where nodes are documents and edges are citation links between documents. The goal of the node classification task is to predict the class of the documents (nodes). The Cora dataset consists of 2,708 nodes and 5,429 edges with 7 classes. Also, the Citesser dataset consists of 3,327 nodes and 4,732 edges with 6 classes. Node features for each dataset consist of bag-of-words for each document. 
As the initial edge features are not given, we set them by concatenating the features of two endpoints of the edge.
We use the classification accuracy as an evaluation metric.

%%%%%%%%%%%%%%%%%%%%%%%%%%%%%%%%%%%%%%%%%%%%%%%%%%%%%%%%%%%%
\begin{figure*}[!t]
    \centering
    \caption{\small \textbf{Additional edge reconstruction results with TopKPool} on the ZINC dataset by varying the compression ratio. Along with the results of Figure 3 in the main paper, we additionally report the average performance of the baselines using TopKPool over 5 different runs with the standard deviation.
    }
    \vspace{-0.05in}
    \includegraphics[width=0.8\linewidth]{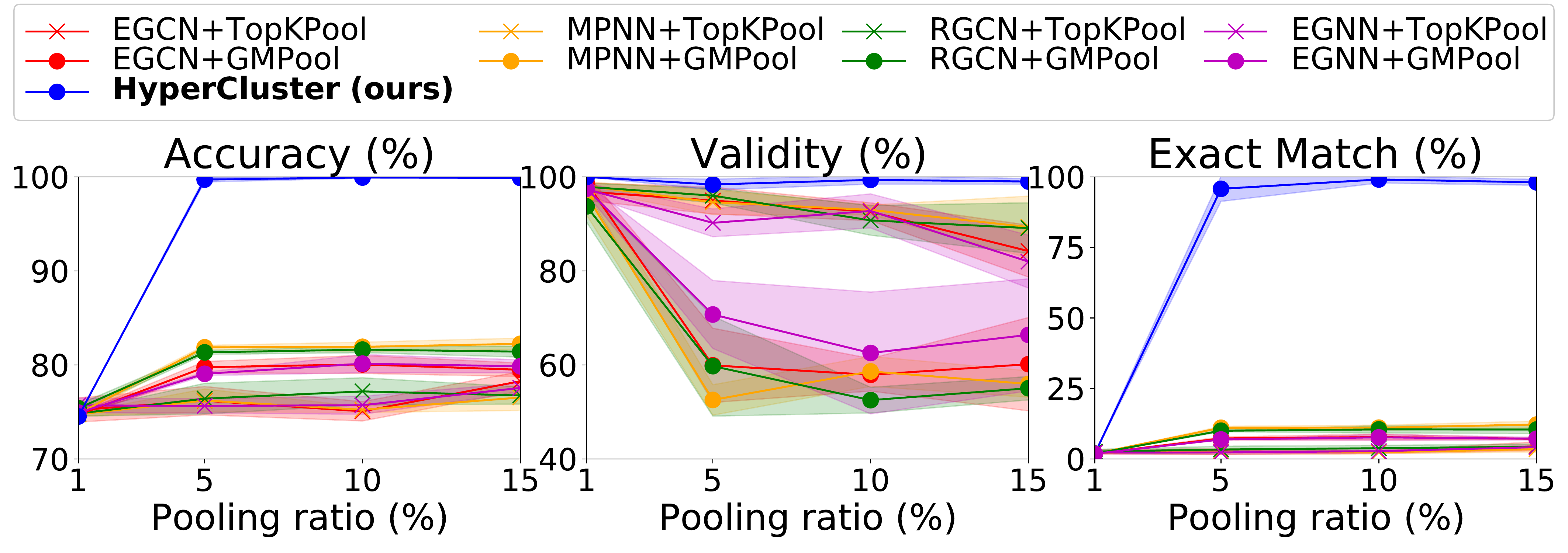}
    \label{fig:recon_edge_addtional}
    % \vspace{-0.05in}
    \includegraphics[width=0.8\linewidth]{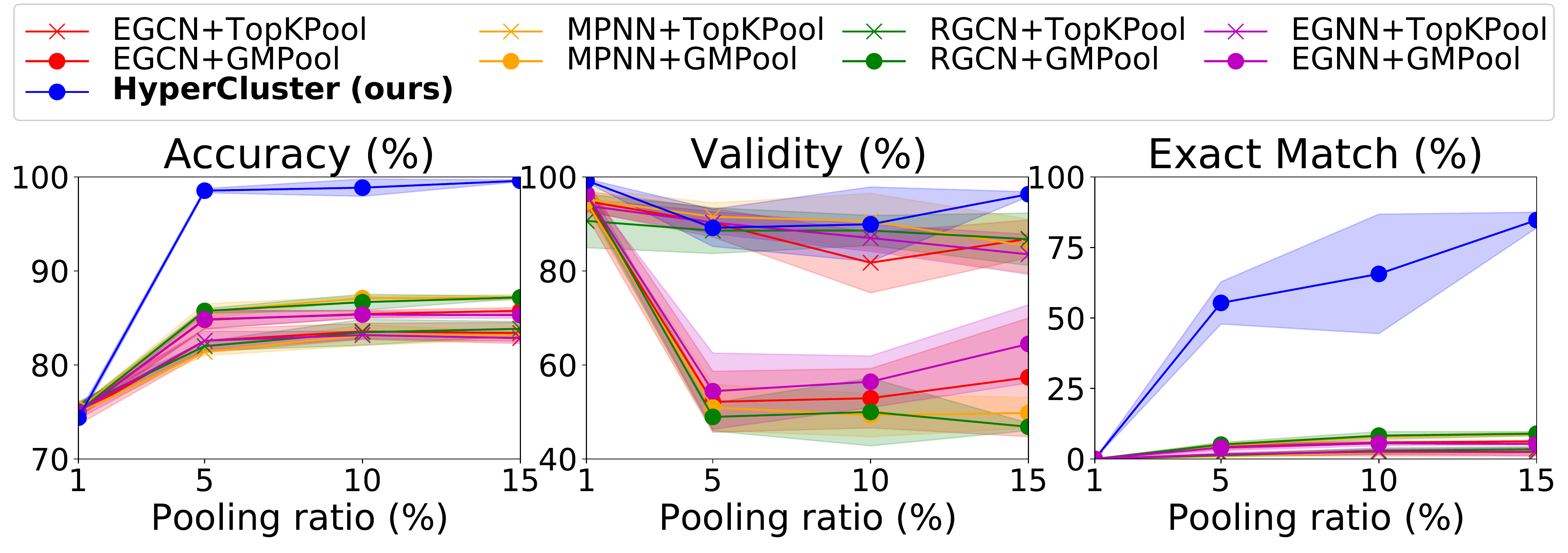}
    \vspace{-0.02in}
    \caption{\small \textbf{Additional graph reconstruction results with TopKPool} on the ZINC dataset by varying the compression ratio. Along with the results of Figure 5 in the main paper, we additionally report the average performance of the baselines using TopKPool over 5 different runs with the standard deviation.
    }
    \label{fig:recon_total_addtional}
    \vspace{-0.1in}
\end{figure*}
%%%%%%%%%%%%%%%%%%%%%%%%%%%%%%%%%%%%%%%%%%%%%%%%%%%%%%%%%%%%

\vspace{-0.05in}
\paragraph{Implementation details}
For a fair evaluation of the semi-supervised node classification task, we follow the standard experimental setting of existing works~\cite{GCN, GAT, grand}, from the node features to the dataset splitting. 
Regarding baselines, we use the naive GCN~\cite{GCN}, GCN with batch normalization~\cite{batchnorm}, and random edge drop scheme~\cite{DropEdge}. Specifically, for the GCN with batch normalization, we use the batch normalization layer between every GCN layer to normalize the features of nodes. Also, for the random edge drop baseline, we randomly drop the partial number of edges before the first layer of GNNs, following the setting of~\citet{DropEdge}, where we do not use the batch normalization to directly see the effect of random drop on the over-smoothing problem. 
For our model, we use the HyperDrop with EHGNN (see section~\ref{sup/pool/drop} for detailed architectures), where we drop edges when passing through every four GNN layers starting from the second layer, and we do not use the batch normalization.
Finally, we use the GCN as the node-level message passing layers for all models, and also use it as the edge-level message passing layers for our HyperDrop with EHGNN.

Following the hyperparameters of the existing semi-supervised node classification work~\cite{grand}, for the Cora dataset, we set the dropout rate as $0.5$, hidden size as $32$, and learning rate as $0.01$. Also, for the Citeseer dataset, we use the same setting from the Cora dataset except for the dropout rate which is set to $0.2$. For the random drop and our models, we drop $20\%$ of edges at each drop step.

\section{Additional Experimental Results \label{sup/result}}
\vspace{-0.05in}
In this section, we provide the additional experimental results on graph reconstruction and generation tasks, with examples of reconstructed or generated molecules. Then, to further qualitatively evaluate the performances of our model, we visualize the edge pooling process of the proposed HyperDrop.

%%%%%%%%%%%%%%%%%%%%%%%%%%%%%%%%%%%%%%%%%%%%%%%%%%%%%%%%%%%%%
\begin{figure*}[!t]
    \centering
    \includegraphics[width=0.7\linewidth]{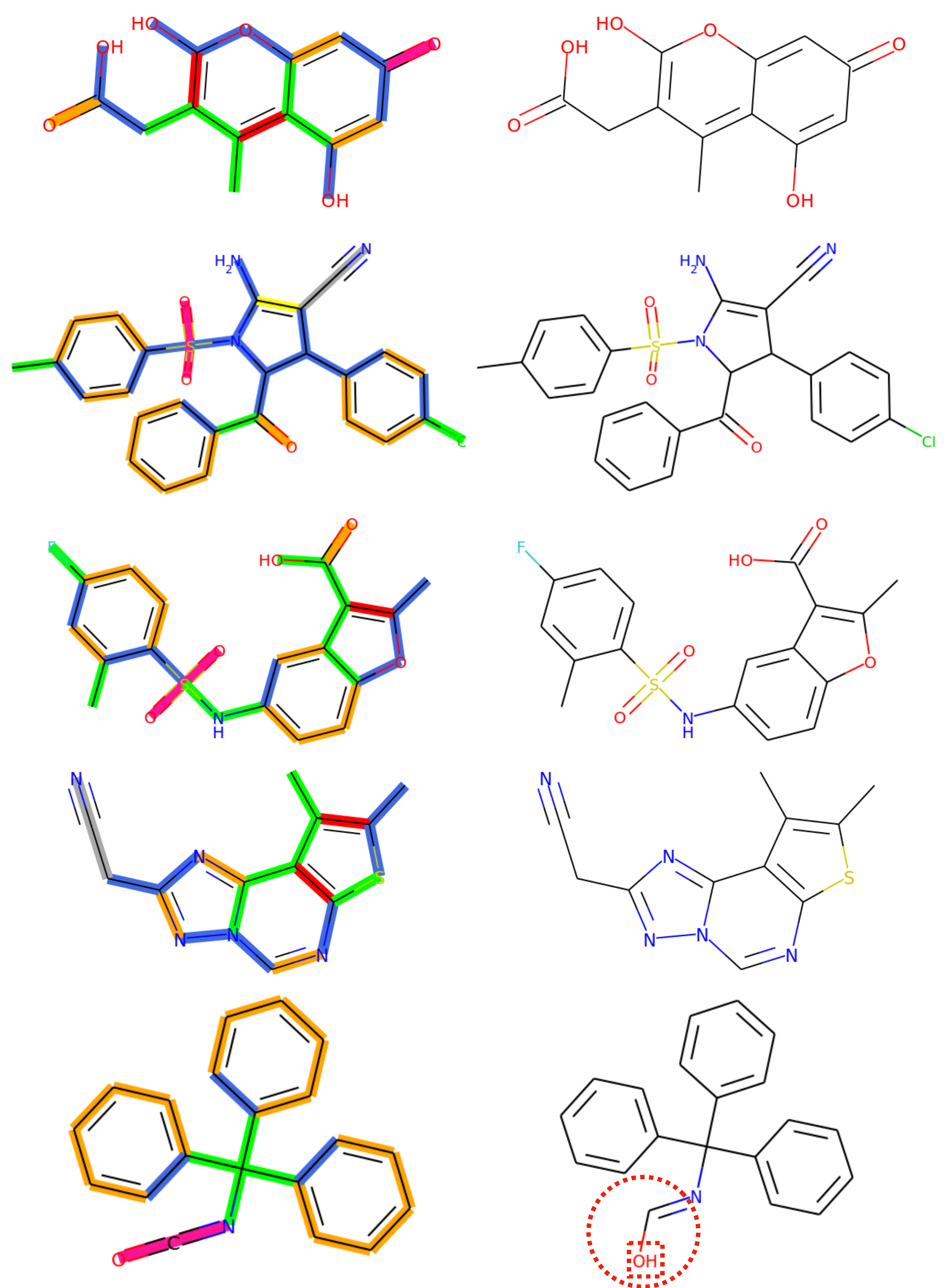}
    % \vspace{-0.08in}
    \caption{\small \textbf{Molecule reconstruction examples.} Molecules shown in the left column are the original molecules with an assigned cluster on each edge, where each cluster is represented as color. The clusters are generated by our method, HyperCluster. The molecules shown in the right column are the reconstructed molecules with our method, where red circles and squares indicate the incorrect prediction of edges and nodes, respectively.
    }
    \label{fig:mol_additional}
    \vspace{-0.2in}
\end{figure*}
%%%%%%%%%%%%%%%%%%%%%%%%%%%%%%%%%%%%%%%%%%%%%%%%%%%%%%%%%%%%%

\subsection{Graph reconstruction \label{sup/result/recon}}
\vspace{-0.05in}

\paragraph{Additional graph reconstruction results}
To see the effect of the pooling method on edge and graph reconstruction tasks, we additionally provide the performance of the TopKPool, a representative node drop method, with existing edge-aware GNN baselines as well as the performance of the GMPool, a node clustering method used in our main paper. 
For the comparison of the pooling methods, we report the performances of both TopKPool and GMPool, in Figure~\ref{fig:recon_edge_addtional} for edge reconstruction and in Figure~\ref{fig:recon_total_addtional} for graph reconstruction. 
As shown in Figure~\ref{fig:recon_edge_addtional} and Figure~\ref{fig:recon_total_addtional}, the proposed EHGNN with HyperCluster largely outperforms all the baselines, which suggests that accurately learning the edge representations is more important than choosing which pooling methods to use, in order to obtain the global graph-level representations. 
Moreover, we observe that the node drop method (TopKPool) for reconstruction is inferior to the node clustering method (GMPool) in terms of accuracy and exact match, since drop methods result in the removal of nodes and edges. The performance gain in validity with the TopKPool mostly comes from its reconstruction of a graph with a single bond, which makes them valid but far different from the desired reconstructed molecules. 

\paragraph{Additional examples of molecular graph reconstruction}
We provide additional examples of reconstructed molecular graphs on the ZINC dataset in Figure~\ref{fig:mol_additional}. Molecules on the left side are the original molecules with each edge color indicating the assigned cluster, obtained by our HyperCluster. Molecules on the right side are the reconstructed molecules, where red circles and squares denote the incorrect predictions of edges and nodes, respectively. As shown in Figure~\ref{fig:mol_additional}, we can see that the clusters are meaningfully assigned with respect to the underlying substructures considering both edges and nodes. For example, edges in the hexagonal ring are assigned to orange and blue colors, where their color patterns are generally determined by the number of adjacent edges with their bond type. Moreover, triple bonds connected to the nitrogen (N) are assigned to the silver-colored cluster.

%%%%%%%%%%%%%%%%%%%%%%%%%%%%%%%%%%%%%%%%%%%%%%%%%%%%%%%%
\begin{figure}[!t]
    \caption{\small \textbf{Graph generation results on MolGAN.} Along with the results of Figure 8 in the main paper, we additionally report the performance of the combination of MolGAN, EHGNN, and GMPool. Solid lines denote the mean, and shaded areas denote the standard deviation of 3 different runs.}
    \centering
    \vspace{-0.1in}
    \begin{subfigure}{0.46\textwidth}
        \includegraphics[width=\linewidth]{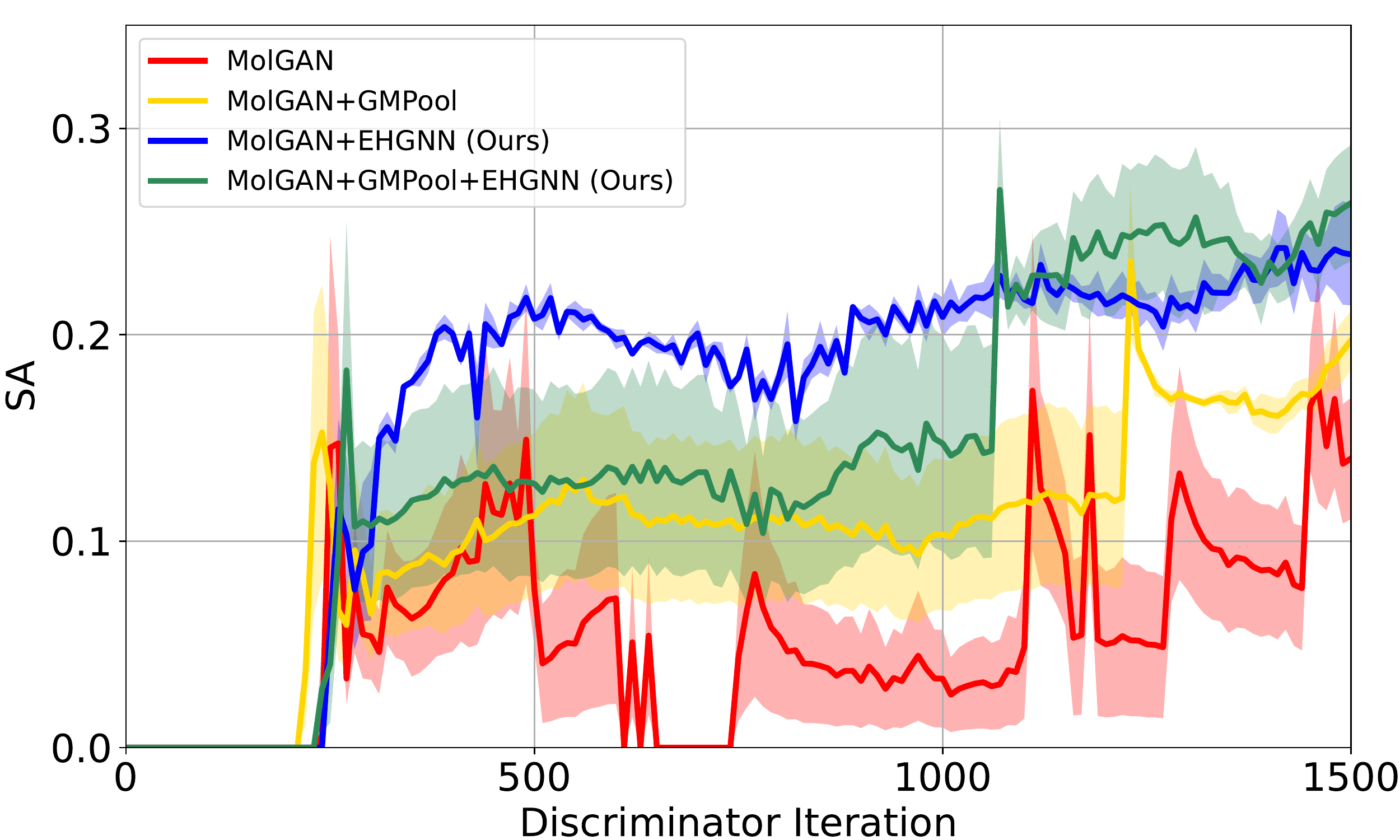}
        \label{fig:molgan_sa_supp}
    \end{subfigure}%
    % \hspace*{\fill}
    \begin{subfigure}{0.46\textwidth}
        \includegraphics[width=\linewidth]{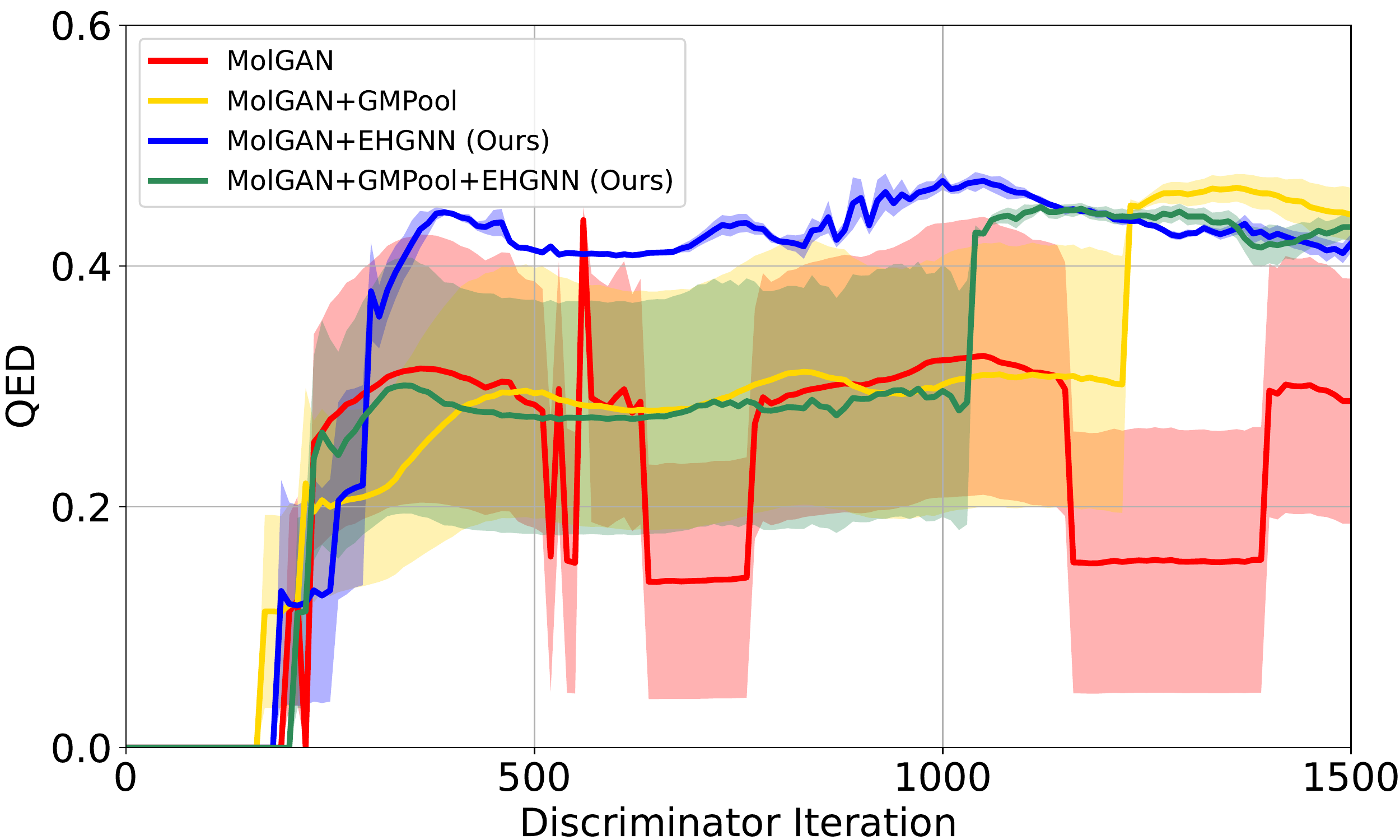}
        \label{fig:molgan_qed_supp}
    \end{subfigure}
    \label{fig:molgan_supp}
    \vspace{-0.2in}
\end{figure}
%%%%%%%%%%%%%%%%%%%%%%%%%%%%%%%%%%%%%%%%%%%%%%%%%%%%%%%%

\begin{figure}[t!]
    \begin{minipage}{0.49\linewidth}
        \centering
        \resizebox{\linewidth}{!}{
        \renewcommand{\arraystretch}{1.0}
        \renewcommand{\tabcolsep}{0.9mm}
        \begin{tabular}{lcccc}
    \toprule
            \textbf{Datasets} & \textbf{Metrics} &\textbf{MARS} & \textbf{MARS + EHGNN (Ours)} \\
        \midrule
            \multirow{5}{*}{ZINC15} & Success Rate & 59.53 $\pm$ 2.11 & \textbf{64.30} $\pm$ 1.54 \\
            & QED ($\ge$ 0.67) & 95.71 $\pm$ 0.09 & \textbf{96.36} $\pm$ 0.49 \\
            & SA ($\ge$ 0.67) & \textbf{99.99} $\pm$ 0.01 & \textbf{99.99} $\pm$ 0.02 \\
            & GSK3$\beta$ ($\ge$ 0.6) & 86.52 $\pm$ 1.67 & \textbf{90.63} $\pm$ 2.57 \\
            & JNK3 ($\ge$ 0.6) & 71.52 $\pm$ 4.15 & \textbf{73.60} $\pm$ 1.29  \\
        \midrule
            \multirow{5}{*}{ChEMBL} & Success Rate & 56.64 $\pm$ 5.79 & \textbf{58.25} $\pm$ 6.07 \\
            & QED ($\ge$ 0.67) & 91.01 $\pm$ 2.79 & \textbf{91.13} $\pm$ 4.84 \\
            & SA ($\ge$ 0.67) & 99.99 $\pm$ 0.01 & \textbf{100.00} $\pm$ 0.00 \\
            & GSK3$\beta$ ($\ge$ 0.6) & 87.45 $\pm$ 1.73 & \textbf{90.34} $\pm$ 2.65  \\
            & JNK3 ($\ge$ 0.6) & \textbf{70.57} $\pm$ 4.75 & 70.01 $\pm$ 4.83  \\
        \bottomrule
    \end{tabular}}
    \vspace{-0.05in}
    \captionof{table}{\small \textbf{Graph generation results on MARS including all evaluation metrics.} The results are the mean and standard deviation of 3 runs.}
    \label{tab:MARS_modified_threshold}
    \end{minipage}
    \hfill
    \begin{minipage}{0.49\linewidth}
        \centering
        \resizebox{\linewidth}{!}{
        \renewcommand{\arraystretch}{1.0}
        \renewcommand{\tabcolsep}{0.9mm}
        \begin{tabular}{lcccc}
        \toprule
        \textbf{Datasets} & \textbf{Metrics} & \textbf{MARS} & \textbf{MARS + EHGNN (Ours)} \\
    \midrule
        \multirow{5}{*}{ZINC15} & Success Rate & 95.65 $\pm$ 0.90 & \textbf{97.28} $\pm$ 1.14 \\
        & QED ($\ge$ 0.6) & 99.07 $\pm$ 0.29 & \textbf{99.45} $\pm$ 0.15 \\
        & SA ($\ge$ 0.67) & \textbf{99.99} $\pm$ 0.01 & \textbf{99.99} $\pm$ 0.02 \\
        & GSK3$\beta$ ($\ge$ 0.5) & 99.13 $\pm$ 0.12 & \textbf{99.52} $\pm$ 0.23 \\
        & JNK3 ($\ge$ 0.5) & 97.33 $\pm$ 1.30 & \textbf{98.21} $\pm$ 0.89 \\
    \midrule
        \multirow{5}{*}{ChEMBL} & Success Rate & \textbf{92.03} $\pm$ 3.83 & 91.88 $\pm$ 3.50 \\
        & QED ($\ge$ 0.6) & \textbf{96.76} $\pm$ 1.44 & 96.43 $\pm$ 2.83 \\
        & SA ($\ge$ 0.67) & 99.99 $\pm$ 0.01 & \textbf{100.00} $\pm$ 0.00 \\
        & GSK3$\beta$ ($\ge$ 0.5) & 99.19 $\pm$ 0.31 & \textbf{99.39} $\pm$ 0.23 \\
        & JNK3 ($\ge$ 0.5) & 95.83 $\pm$ 2.30 & \textbf{95.85} $\pm$ 0.92 \\
    \bottomrule
    \end{tabular}}
    \vspace{-0.05in}
    \captionof{table}{\small \textbf{Graph generation results on MARS under the setting of original success thresholds.} The results are the mean and standard deviation of 3 runs.}
    \label{tab:MARS_original_threshold}
    \end{minipage}
    \vspace{-0.1in}
\end{figure}
%%%%%%%%%%%%%%%%%%%%%%%%%%%%%%%%%%%%%%%%%%%%%%%%%%%%%%%%%%%%%

\subsection{Graph generation \label{sup/result/gen}}

\paragraph{MolGAN}
Since the EHGNN framework can be jointly used with the node-level representation learning methods, we can further combine the EHGNN framework with the node pooling method, for obtaining holistic graph-level representation from both node and edge representations. Thus, we additionally couple the MolGAN + EHGNN with the state-of-the-art node pooling method, namely GMPool. As shown in Figure~\ref{fig:molgan_supp}, compared to the large performance gain obtained by our EHGNN, the performance gain obtained from using both GMPool and EHGNN is relatively small, and also the training using both architectures is unstable. This might be because, we can already obtain the effective graph-level representation only with the combination of MolGAN and EHGNN, and additionally using more layers makes the training of the MolGAN architecture difficult since this scheme also increases the number of parameters. On the other perspective, since the original MolGAN architecture is already able to utilize the node representations, albeit, by simple R-GCN, the remaining performance gain comes from the explicit edge representations via our EHGNN.

\paragraph{MARS}
Here, we provide the additional experimental results using the MARS architecture on the ChEMBL dataset, where we used the available data\footnote{\label{footnote:mars}https://github.com/yutxie/mars} from~\citet{MARS}. As shown in Table~\ref{tab:MARS_modified_threshold}, MARS equipped with our EHGNN outperforms the baseline model, showing the same tendency as in the results on the ZINC15 dataset.
Also, the original MARS and the MARS with EHGNN models successfully generate the high-quality molecules in terms of SA, and there is not much significant difference between those two models on this metric. However, the performance gain with our EHGNN against the naive MARS comes from other metrics, such as QED and GNK3$\beta$, resulting in the successful generation of molecules having all desired properties. 

On the other hand, we also report the success rate with individual evaluation metrics according to thresholds used in the MARS paper~\cite{MARS} in Table~\ref{tab:MARS_original_threshold}. As shown in Table~\ref{tab:MARS_original_threshold}, our MARS + EHGNN model still outperforms the baseline on most of the metrics, and the performance tendency is highly similar to the result of different thresholds in Table~\ref{tab:MARS_modified_threshold}. Those two results demonstrate that accurate learning of edge representation is important to generate desirable molecules.

%%%%%%%%%%%%%%%%%%%%%%%%%%%%%%%%%%%%%%%%%%%%%%%%%%%%%%%%%%%%%
\begin{figure*}[!t]
    \centering
    \includegraphics[width=0.9\linewidth]{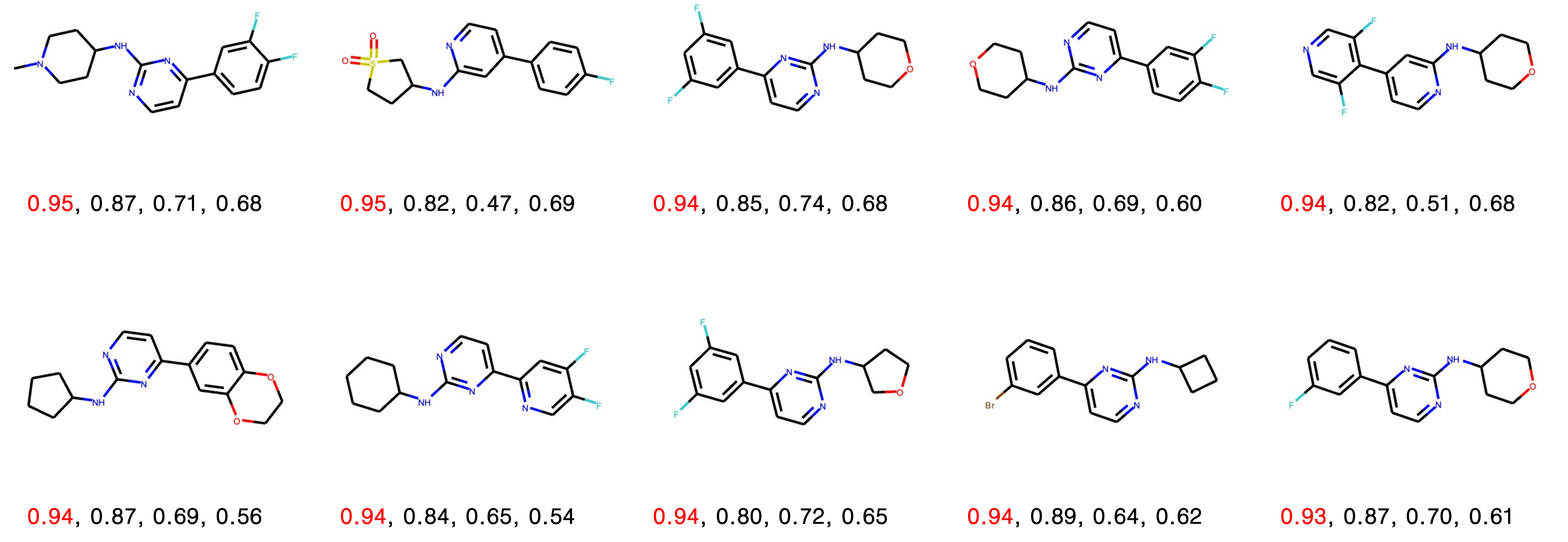}
    \vspace{-0.075in}
    \caption{\small \textbf{10 generated molecules with the highest QED scores.} The numbers are QED, SA, GSK3$\beta$, and JNK3 scores, respectively. We highlight the QED score in red among four different scores.
    }
    \label{fig:mars_qed}
    \vspace{0.1in}
    \includegraphics[width=0.9\linewidth]{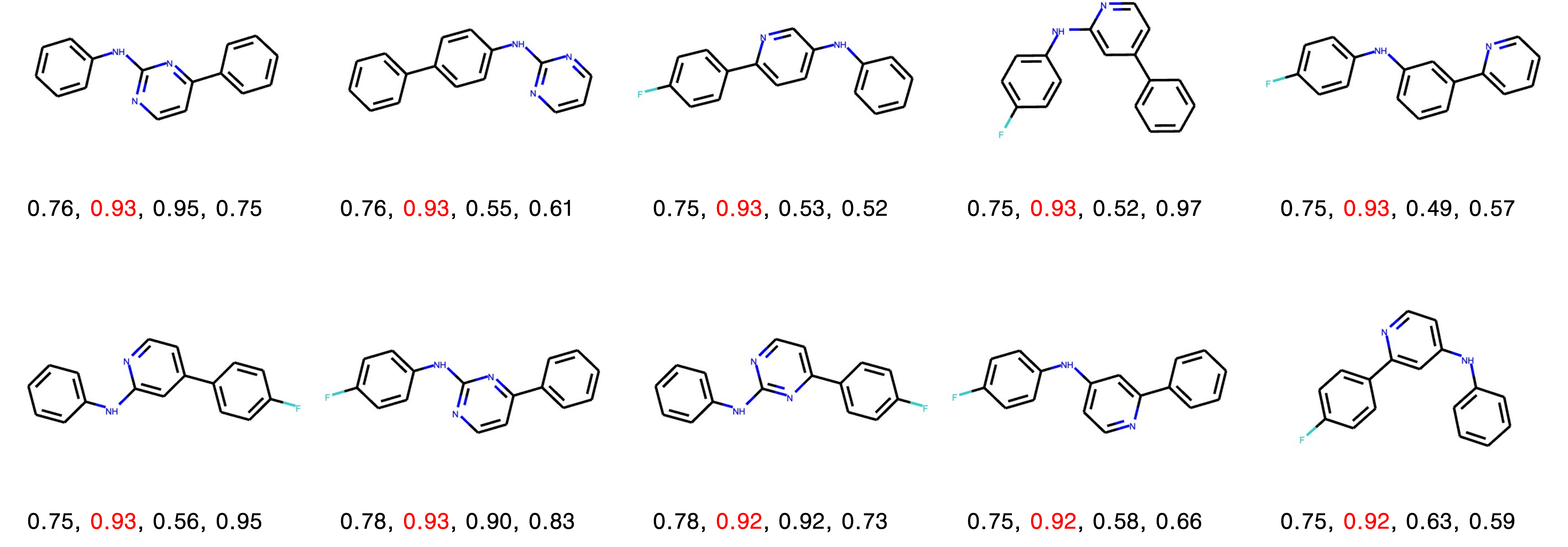}
    \vspace{-0.075in}
    \caption{\small \textbf{10 generated molecules with the highest SA scores.} The numbers are QED, SA, GSK3$\beta$, and JNK3 scores, respectively. We highlight the SA score in red among four different scores.
    }
    \label{fig:mars_sa}
    \vspace{0.1in}
    \includegraphics[width=0.9\linewidth]{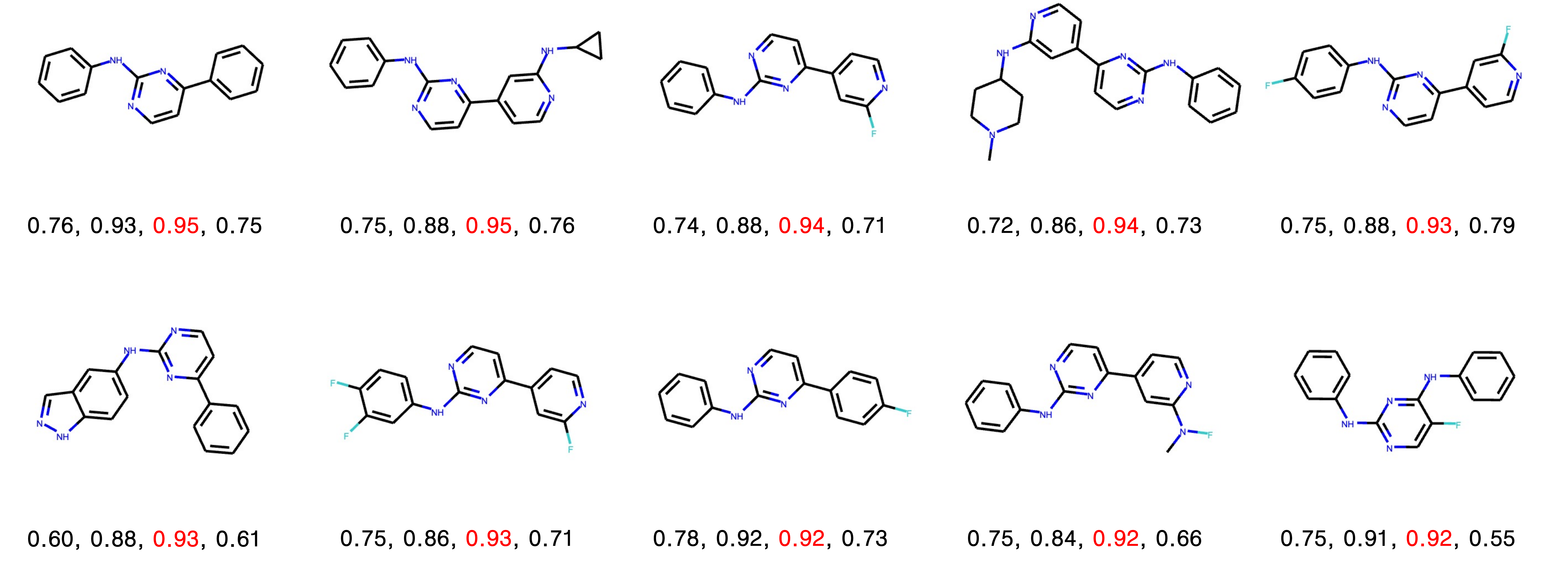}
    \vspace{-0.075in}
    \caption{\small \textbf{10 generated molecules with the highest GSK3$\beta$ scores.} The numbers are QED, SA, GSK3$\beta$, and JNK3 scores, respectively. We highlight the GSK3$\beta$ score in red among four different scores.
    }
    \label{fig:mars_gsk3b}
    \vspace{0.1in}
    \includegraphics[width=0.9\linewidth]{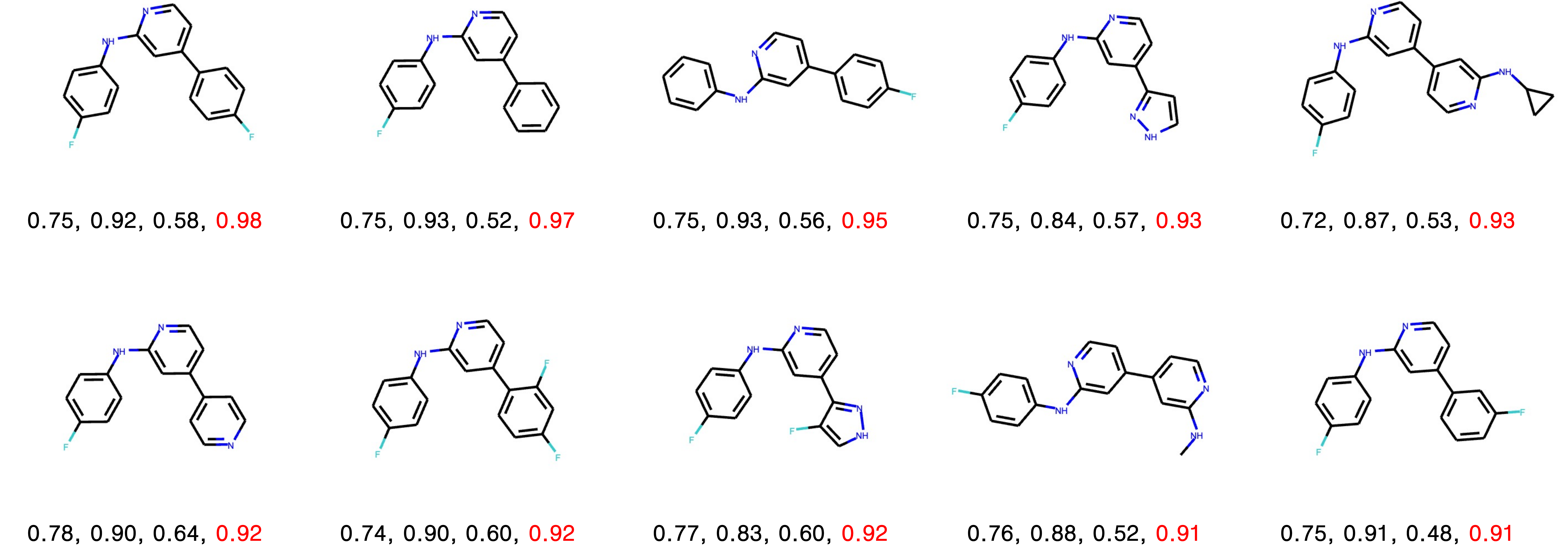}
    \vspace{-0.075in}
    \caption{\small \textbf{10 generated molecules with the highest JNK3 scores.} The numbers are QED, SA, GSK3$\beta$, and JNK3 scores, respectively. We highlight the JNK3 score in red among four different scores.
    }
    \label{fig:mars_jnk3}
    \vspace{-0.15in}
\end{figure*}
%%%%%%%%%%%%%%%%%%%%%%%%%%%%%%%%%%%%%%%%%%%%%%%%%%%%%%%%%%%%%

\paragraph{Visualization of the generated molecular graphs} 
We further provide the examples of generated molecules using our EHGNN on MARS in Figure~\ref{fig:mars_qed}, \ref{fig:mars_sa}, \ref{fig:mars_gsk3b}, and \ref{fig:mars_jnk3}. We hope that these examples are to be helpful for the chemists to get an insight into the molecules generated with our framework.

%%%%%%%%%%%%%%%%%%%%%%%%%%%%%%%%%%%%%%%%%%%%%%%%%%%%%%%%%%%%

\begin{figure*}[!t]
    \centering
    \includegraphics[width=0.9\linewidth]{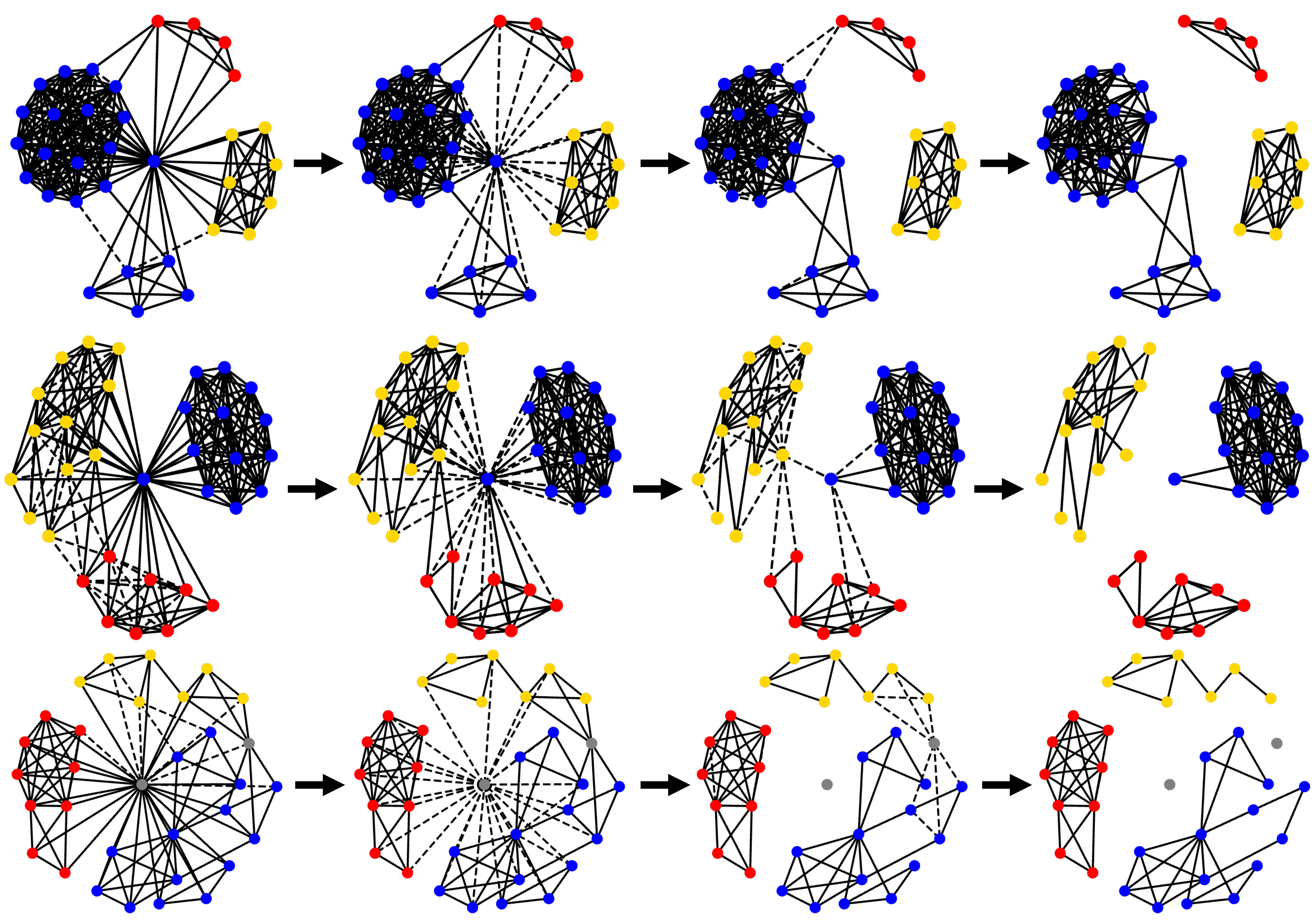}
    % \vspace{-0.08in}
    \caption{\small \textbf{Edge pooling results on the COLLAB dataset.} Each row represents the pooling process of a graph. Colors denote connected components.
    }
    \label{fig:collab_supp}
    \includegraphics[width=0.9\linewidth]{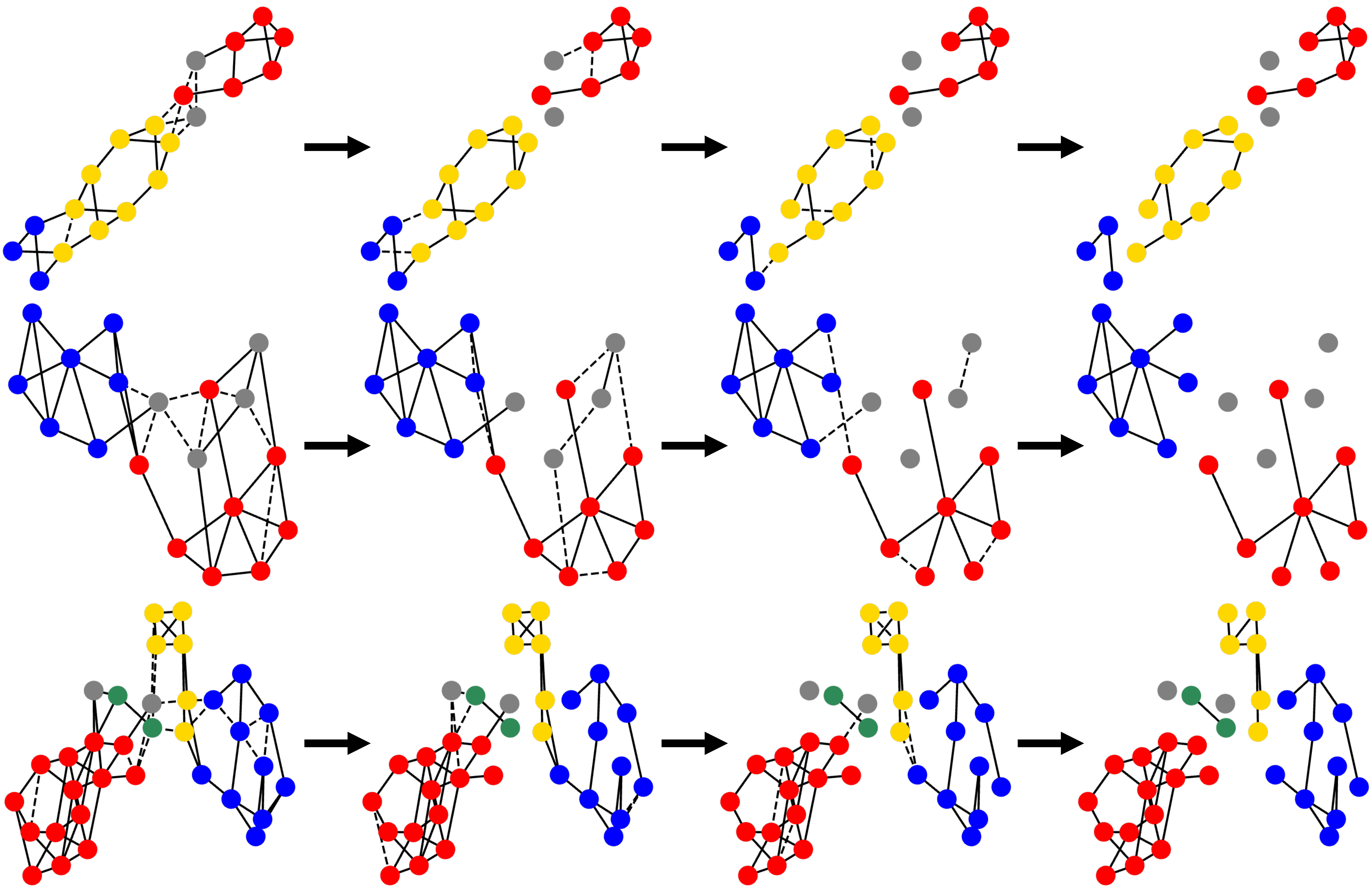}
    % \vspace{-0.08in}
    \caption{\small \textbf{Edge pooling results on the PROTEINS dataset.} Each row represents the pooling process of a graph. Colors denote connected components.
    }
    \label{fig:proteins_supp}
    \vspace{-0.16in}
\end{figure*}

%%%%%%%%%%%%%%%%%%%%%%%%%%%%%%%%%%%%%%%%%%%%%%%%%%%%%%%%%%%%

\newpage

\subsection{Graph classification \label{sup/result/class}}
\paragraph{Additional examples of HyperDrop process}
We provide additional examples of HyperDrop processes on the COLLAB and PROTEINS datasets in Figure~\ref{fig:collab_supp} and Figure~\ref{fig:proteins_supp}, respectively. Colors represent the resulting connected components in the final graph after dropping edges, and we represent isolated nodes as gray. Arrows indicate the layer-wise progressive pooling processes. We can see that by dropping unnecessary edges, a large graph is divided into smaller connected components, which we assume to be effective for message passing between the relevant nodes.

\section{Limitations and Potential Societal Impacts \label{sup/impact}}
In this section, we discuss the limitations and potential societal impacts of our work.

\paragraph{Limitations}

In this work, we propose to learn edge representations with hypergraphs, using the dual hypergraph transformation that allows us to apply off-the-shelf node-level message-passing schemes designed for node representation learning to edges. While we can learn accurate edge representations using the proposed framework, we need two separate GNNs to learn node and edge representations independently. Combining these two GNNs into one, by learning node and edge representations jointly using a single GNN, may be more effective for learning graph representations, while saving the memory as well. We leave this as future work.

\paragraph{Potential societal impacts}

The system for generating target molecules is significantly important to our society, since it can be used to generate vaccines or drugs for diseases, even for the newly emerged severe acute respiratory syndrome coronavirus 2 (SARS-CoV-2). However, the conventional development of beneficial molecules requires a huge amount of time and resources with a significant number of trial-and-error processes, before actually applying the generated molecules, since we have to check potential outcomes those molecules can have.

In this paper, we show that the proposed edge representation learning framework can accurately represent the edges of the given graph, for the holistic graph-level representation learning, which has been extensively validated on graph generation and classification tasks with biochemical molecules. Therefore, this approach can meaningfully aid the development of target molecules in the following ways. First, the generation system described in Section 4.2 of the main paper is effective for generating molecules with desirable properties, since it can generate more drug-like molecules that can effectively inhibit multiple target proteins. Also, the classification system described in Section 4.3 of the main paper is beneficial for examining the toxicity of generated molecules, which is an essential step before human clinical trials or being deployed on a commercial scale. Therefore, our method allows us to reduce time and resources for generating and validating target molecules, for example in the domain of de novo drug design compared to synthesizing drugs by trial-and-error.

As described above, while our method has huge potential impacts for discovering novel molecules in our real-life, anyone can maliciously use our system, aiming to develop harmful compounds for humans, such as synthesizing toxic or addictive substances. Thus, we strongly hope that our method would not be applied for generating harmful molecules that may have negative impacts on our society.

\end{document}